\documentclass[lettersize,journal]{IEEEtran}
\usepackage{amsmath,amsfonts}
\usepackage{algorithmic}
\usepackage{algorithm}
\usepackage{array}
\usepackage[caption=false,font=normalsize,labelfont=sf,textfont=sf]{subfig}
\usepackage{textcomp}
\usepackage{stfloats}
\usepackage{url}
\usepackage{verbatim}
\usepackage{graphicx}
\usepackage{cite}
\newcommand{\modelname}{LegoOcc}
\newcommand{\newmodelname}{LegoOcc++}

\newcommand{\clsname}[2]{ \multicolumn{1}{c}{ \rotatebox{90}{ \hspace{-6pt} \textcolor{#2}{$\blacksquare$}
#1 }}}

\newcommand{\graycell}[1]{\textcolor{gray}{#1}}

\newcommand{\ie}{i.e.,\xspace}

\usepackage{xspace}
\providecommand{\ie}{\emph{i.e.}\xspace}

\usepackage[table]{xcolor}
\definecolor{impblue}{RGB}{64,150,230}

\usepackage{multirow}

\usepackage{colortbl}

\usepackage{booktabs}

\usepackage{bm}

\usepackage{graphicx}

\usepackage{caption}

\usepackage{array}

\usepackage{amssymb}

\usepackage{amsmath}

\usepackage{wrapfig}

\usepackage{placeins}

\usepackage{booktabs,makecell,multirow,pifont}

\definecolor{ceiling}{RGB}{214,  38, 40}   %
\definecolor{floor}{RGB}{43, 160, 4}     %
\definecolor{wall}{RGB}{158, 216, 229}  %
\definecolor{window}{RGB}{114, 158, 206}  %
\definecolor{chair}{RGB}{204, 204, 91}   %
\definecolor{bed}{RGB}{255, 186, 119}  %
\definecolor{sofa}{RGB}{147, 102, 188}  %
\definecolor{table}{RGB}{30, 119, 181}   %
\definecolor{tvs}{RGB}{160, 188, 33}   %
\definecolor{furniture}{RGB}{255, 127, 12}  %
\definecolor{objects}{RGB}{196, 175, 214} %

\definecolor{barrier}{RGB}{112, 128, 144}
\definecolor{bicycle}{RGB}{220, 20, 60}
\definecolor{bus}{RGB}{255, 127, 80}
\definecolor{car}{RGB}{255, 158, 0}
\definecolor{construct}{RGB}{233, 150, 70}
\definecolor{motor}{RGB}{255, 61, 99}
\definecolor{pedestrian}{RGB}{0, 0, 230}
\definecolor{traffic}{RGB}{47, 79, 79}
\definecolor{trailer}{RGB}{255, 140, 0}
\definecolor{truck}{RGB}{255, 98, 70}
\definecolor{driveable}{RGB}{0, 207, 191}
\definecolor{other}{RGB}{175, 0, 75}
\definecolor{sidewalk}{RGB}{75, 0, 75}
\definecolor{terrain}{RGB}{112, 180, 60}
\definecolor{manmade}{RGB}{222, 184, 135}
\definecolor{vegetation}{RGB}{0, 175, 0}
\definecolor{others}{RGB}{0, 0, 0}

\definecolor{Sbg}{HTML}{F4F9FF}        %
\definecolor{Ubg}{HTML}{FFF7F0}        %
\definecolor{SbgStrong}{HTML}{DCEBFF}  %
\definecolor{UbgStrong}{HTML}{FFE2CC}  %
\definecolor{SbgVivid}{HTML}{91b4ed} 
\definecolor{UbgVivid}{HTML}{f2aa61}

\usepackage[hidelinks]{hyperref}
\usepackage{cleveref}

\begin{document}

\title{LegoOcc++: Towards Consistent Open-Vocabulary 3D Occupancy Prediction}

\author{Changqing Zhou,
Yueru Luo,
Han Zhang,
Zeyu Jiang,
and Changhao Chen%
\thanks{Changqing Zhou, Han Zhang, Zeyu Jiang and Changhao Chen are with The Hong Kong University
of Science and Technology (Guangzhou), Guangzhou 511453, China (e-mail: czhou149@connect.hkust-gz.edu.cn; changhaochen@hkust-gz.edu.cn). Changhao Chen is also with the Division of Emerging Interdisciplinary Areas, The Hong Kong University of Science and Technology, Hong Kong SAR 999077, China.}%
\thanks{Yueru Luo is with The Chinese University of Hong Kong, Shenzhen, China.}%
\thanks{Corresponding author: Changhao Chen.}%
}

\markboth{Journal of \LaTeX\ Class Files,~Vol.~X, No.~X, September~2026}%
{Shell \MakeLowercase{\textit{et al.}}: A Sample Article Using IEEEtran.cls for IEEE Journals}

\maketitle

\begin{abstract}
Open-vocabulary 3D occupancy prediction is important for embodied agents operating in complex environments whose semantic categories extend beyond fixed taxonomies. While recent studies have explored open-vocabulary occupancy in outdoor driving scenarios, indoor environments remain challenging due to dense occlusions, complex geometry, and fine-grained long-tailed semantics. To reduce the need for expensive voxel-level semantic annotations, we study a geometry-only supervision setting requiring only binary occupancy labels.
We build upon Language-Embedded Gaussians as a unified representation coupling 3D geometry with language-aligned semantics. For geometry learning, we introduce an opacity-aware Poisson-based Gaussian-to-Occupancy formulation that enables stable volumetric aggregation under binary occupancy supervision. For semantic learning, we render Gaussian features into images and align them with a frozen open-vocabulary model, together with Progressive Temperature Decay to reduce feature mixing during Gaussian splatting.
Building on this framework, we further introduce Instance-Level Semantic Consistency to alleviate fragmented object predictions, and Covisibility-Aware Support-View Selection to provide informative and less redundant multi-view semantic supervision.
We also extend our framework to outdoor driving scenarios to evaluate its generality across substantially different environments.
Extensive experiments on indoor and outdoor benchmarks demonstrate that our method achieves strong open-vocabulary occupancy performance, validating the effectiveness and generality of our framework. Code will be available at \url{https://github.com/JuIvyy/LegoOcc}.
\end{abstract}


\section{Introduction}
\label{sec:intro}

Comprehensive 3D geometric and semantic understanding of surrounding environments is fundamental to embodied agents across applications such as service robotics, drones, and AR/VR systems.
Among different scene representations, \emph{3D semantic occupancy} has emerged as a powerful and compact formulation that jointly models scene geometry, semantics, and free space within a unified volumetric representation, providing rich spatial information for downstream reasoning and decision-making~\cite{volumetrivln,occvla,occllama,chen2025constraint}.
Vision-based occupancy prediction is particularly attractive due to the ubiquity, low cost, and rich appearance cues of cameras.
Recent studies have demonstrated promising performance from both \emph{monocular} inputs~\cite{ISO,embodiedocc,embodiedocc++,yu2025shtocc,OccFormer,VoxFormer,SparseOcc,gpocc,jiang2026freeocc,gpoccpp} and \emph{multi-view} inputs~\cite{surroundocc,gaussianformer,gaussianformer2,gaussrender}, using architectures ranging from dense voxel networks and transformers to Gaussian-based representations.

\begin{figure}[t]
  \centering
  \includegraphics[width=0.5\textwidth]{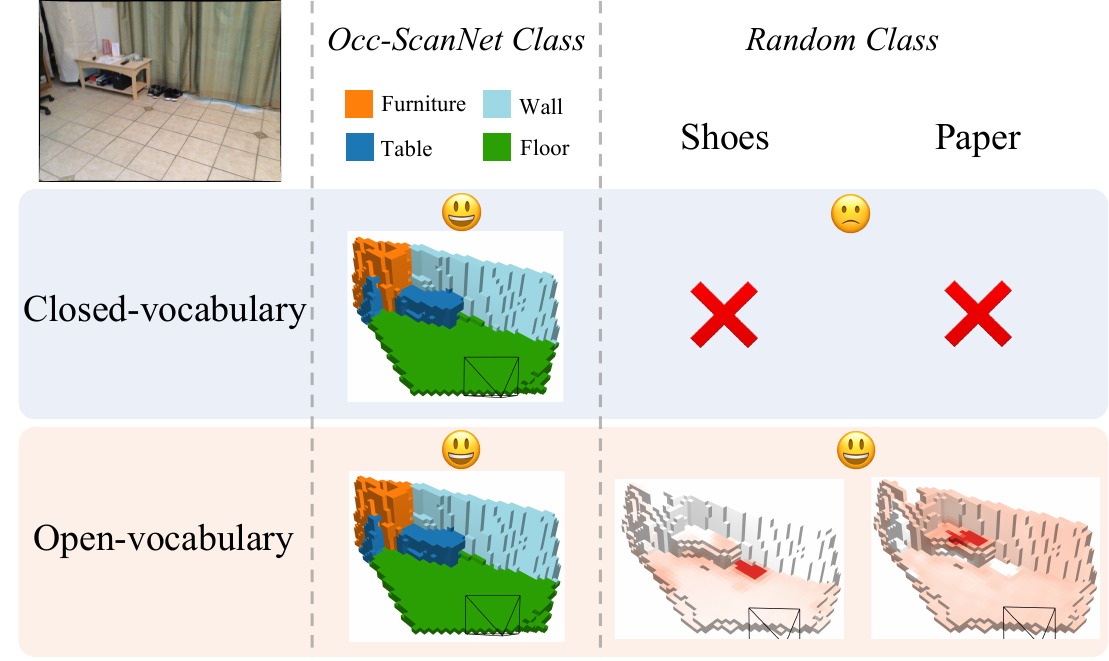}
  \caption{\textbf{Closed- vs.\ open-vocabulary occupancy.}
  Conventional closed-vocabulary methods~\cite{ISO,embodiedocc} can recognize only categories predefined during training.
  In contrast, our open-vocabulary formulation aligns language with 3D occupancy and supports text queries for arbitrary categories.
  \emph{Right column (Random Class):} text-conditioned voxel scores are visualized as heatmaps, where darker red indicates a higher response to the queried category.}
  \label{fig:teaser}
\end{figure}

Despite this progress, most existing occupancy methods operate under a \emph{closed vocabulary}, where semantic categories are predefined during training.
This assumption is restrictive in real-world environments, where category distributions are highly long-tailed~\cite{zhang2023deeplongtailsurvey,chen2025open} and the semantic space is inherently open-ended~\cite{joseph2021towards}.
As illustrated in~\Cref{fig:teaser}, closed-vocabulary models cannot recognize categories beyond their training taxonomy, whereas an open-vocabulary formulation enables arbitrary semantic concepts to be queried directly in 3D space.
Such capability is particularly important for embodied agents, which must recognize and reason about previously unseen objects before interacting with their surroundings.

Open-vocabulary occupancy prediction has recently received increasing attention in autonomous-driving scenarios, where vision-language models and text-aligned representations are incorporated into camera-based 3D occupancy to support free-form semantic reasoning~\cite{veon,yu2025language,boeder2024langocc,li2025ago,groundingocc}.
In contrast, indoor occupancy remains substantially less explored and is typically trained and evaluated under small, predefined taxonomies.
For example, Occ-ScanNet~\cite{ISO} defines only 11 semantic classes, which substantially underrepresent the diversity of real indoor environments.
Indoor scenes further pose two particular challenges: they contain much denser and more heavily occluded geometry, while exhibiting a substantially richer set of fine-grained and long-tailed object categories.
Consequently, directly transferring existing outdoor open-vocabulary occupancy pipelines to indoor environments leads to limited performance~\cite{embodiedocc}.
Early attempts such as OVO~\cite{ovo} distill 2D open-vocabulary segmentation into 3D voxels, but still rely on a predefined set of base categories during training.
These observations motivate a more scalable formulation for indoor open-vocabulary occupancy prediction.

A central challenge is therefore the availability of semantic supervision. Dense voxel-wise semantic annotation is particularly expensive for indoor scenes because it requires labeling a large number of fine-grained and long-tailed categories. In contrast, scene geometry can be obtained much more readily from RGB-D observations and existing reconstruction pipelines. For example, ScanNet provides RGB-D observations, while Occ-ScanNet constructs occupancy ground truth using SCFusion~\cite{scfusion}, which combines depth-based reconstruction~\cite{supereight}, scene-completion methods~\cite{scan2cad,forknet}, and aligned ShapeNet CAD models~\cite{scan2cad}. This observation suggests an attractive learning paradigm: \emph{geometry can provide reliable 3D supervision, while open-vocabulary semantic knowledge can be transferred from pretrained 2D models}. We therefore consider a geometry-only supervision setting, in which training uses binary voxel occupancy labels but does not require semantic voxel annotations, and ask:
\emph{How can open-vocabulary semantics be reliably transferred from images into a geometrically supervised 3D occupancy representation?}

To address this gap, our previous conference work, \modelname{}~\cite{legoocc}, proposed \emph{Language-Embedded Gaussians} (LE-Gaussians)~\cite{shi2023legaussians,zhou2024featuregs} as a unified intermediate representation for geometry and open-vocabulary semantics.
Each LE-Gaussian jointly encodes geometric attributes, including position, covariance, and opacity, together with a language-aligned feature anchored at a specific 3D location.
This representation allows geometry and semantics to be learned through two complementary supervision paths.
For geometry learning, \modelname{} converts continuous Gaussian primitives into voxel occupancy using an opacity-aware \emph{Poisson-based Gaussian-to-Occupancy} formulation.
Unlike existing Gaussian-to-Occupancy mappings that aggregate spatial kernels independently of opacity, our formulation explicitly incorporates Gaussian opacity into volumetric aggregation, making 3D occupancy inference consistent with the opacity-aware rendering used for semantic learning.
For semantic learning, LE-Gaussian features are differentiably rendered into the image plane through Gaussian splatting~\cite{gaussiansplatting} and aligned with features extracted from a frozen training-free open-vocabulary segmenter, such as Trident~\cite{trident}.
Since standard $\alpha$-blending mixes embeddings from multiple Gaussians along each ray, \modelname{} further introduces \emph{Progressive Temperature Decay}, which progressively sharpens Gaussian opacities during training to reduce feature mixing while preserving differentiable supervision.
Together, these designs enable language-aligned 3D occupancy to be learned without semantic voxel annotations.

While \modelname{} demonstrates the effectiveness of Language-Embedded Gaussians for indoor open-vocabulary occupancy prediction, several limitations remain.
First, \modelname{} performs semantic alignment primarily at the pixel level. Although Progressive Temperature Decay reduces feature mixing along individual rays, pixels belonging to different regions of the same object are optimized independently. As a result, the predicted semantics can become fragmented within a single object, with different parts assigned inconsistent categories.
Second, \modelname{} exploits additional frames for semantic supervision but selects support views without explicitly modeling geometric covisibility. Temporally nearby or visually similar frames may provide redundant observations, whereas distant frames may have insufficient shared geometry. The appropriate temporal interval also varies with video frame rate and sampling frequency, making fixed temporal selection difficult to generalize. %
Third, \modelname{} was developed and evaluated primarily on indoor embodied scenes, leaving its applicability to large-scale outdoor environments with different geometry, sensing configurations, and semantic distributions unexplored~\cite{wei2026onebev,tan2024epmf,hong2024unified}. %
These limitations point to two complementary requirements for a more general open-vocabulary occupancy framework: \emph{semantic consistency should be enforced at the object level, and multi-view supervision should be determined by geometric overlap rather than temporal proximity.}

Motivated by these observations, we further propose \textbf{\newmodelname{}}, a substantially enhanced framework that extends \modelname{} in both semantic representation learning and multi-view supervision, while broadening open-vocabulary occupancy prediction from indoor embodied environments to outdoor autonomous-driving scenarios.
First, we propose \emph{Instance-Level Semantic Consistency}, which complements pixel-wise language alignment with an object-level semantic constraint.
Category-agnostic masks generated by a frozen SAM~\cite{sam} are used to aggregate open-vocabulary teacher features within reliable object regions into instance-level semantic prototypes. Rendered LE-Gaussian features associated with the same instance are then encouraged to agree with this shared representation. This constraint provides a direct mechanism for suppressing intra-object semantic fragmentation, complementing the ray-level feature disambiguation provided by Progressive Temperature Decay.
Second, we introduce \emph{Covisibility-Aware Support-View Selection} for more reliable multi-view semantic supervision.
Rather than relying on a fixed temporal interval, we explicitly measure geometric overlap between candidate views using their visible 3D regions. Views with insufficient overlap or nearly duplicate content are filtered out, after which pose-diverse support views are selected to provide complementary observations.
This strategy improves the quality of cross-view feature alignment and makes more effective use of multi-frame language supervision.

We further extend our framework from indoor embodied scenes to outdoor autonomous-driving environments. Unlike the indoor setting, where binary occupancy labels provide geometry supervision, we follow prior open-vocabulary driving methods and train on outdoor data \textbf{without any occupancy annotations}.
This substantially different supervision setting, together with the large differences in scene scale and semantic distribution, provides a broader test of the effectiveness and generality of our approach. On Occ-ScanNet, \newmodelname{} improves \modelname{} from 59.50 to 61.73 IoU and from 21.05 to 24.22 mIoU. On Occ3D-nuScenes, \newmodelname{} achieves 15.14 mIoU with single-frame inference, outperforming all compared single-frame methods and nearly matching the 15.15 mIoU of the eight-frame PGOcc~\cite{yan2025progressive}. These results demonstrate that our approach generalizes effectively across substantially different scene domains and supervision settings.

Compared with \modelname{}, the additional contributions of this work are summarized as follows:
\begin{itemize}
    \item We propose \textbf{\newmodelname{}}, an enhanced open-vocabulary occupancy framework that extends \modelname{} with object-level semantic consistency and geometry-aware multi-view supervision.

    \item We introduce \textbf{Instance-Level Semantic Consistency}, which leverages category-agnostic SAM masks and open-vocabulary teacher features to construct object-level semantic prototypes and regularize rendered LE-Gaussian features, reducing semantic fragmentation within individual objects.

    \item We introduce \textbf{Covisibility-Aware Support-View Selection}, which explicitly measures shared 3D content between views, filters both insufficiently overlapping and highly redundant observations, and selects pose-diverse support views for more reliable cross-view semantic alignment.

    \item We extend our framework from indoor embodied environments under geometry-only supervision to outdoor autonomous-driving scenarios under annotation-free training, and conduct substantially expanded experiments and analyses to demonstrate its effectiveness and generality across diverse scene domains and supervision settings.
\end{itemize}

This article builds upon our previous conference paper, \modelname{}~\cite{legoocc}, published at CVPR 2026 as an Oral presentation. The present work introduces two new methodological components---Instance-Level Semantic Consistency and Covisibility-Aware Support-View Selection---and substantially expands the scope of the framework to outdoor autonomous-driving scenes and a different supervision regime. Extensive experiments and analyses further investigate the effectiveness, generality, and limitations of open-vocabulary occupancy prediction across indoor and outdoor environments.

\section{Related Works}\label{sec:related_work}

\subsection{Occupancy Prediction}

Camera-only occupancy prediction has been extensively studied in outdoor autonomous-driving scenarios, often using grid-based or transformer-based pipelines~\cite{gaussianformer,gaussianformer2,occupancypoints,li2026hierarchical}, whereas indoor settings remain comparatively underexplored.
MonoScene formulated monocular semantic scene completion from RGB images~\cite{monoscene,sscnet,ndcscene}, and subsequent works on indoor benchmarks such as Occ-ScanNet~\cite{ISO} further improved prediction accuracy and scalability.
Representative methods include ISO~\cite{ISO}, which adopts depth-guided feature lifting; EmbodiedOcc~\cite{embodiedocc}, which employs Gaussian-based volumetric prediction; and RoboOcc~\cite{roboocc}, which introduces opacity-guided encoders.
Methodologically, indoor approaches range from dense volumetric lifting of 2D features along camera rays into voxel grids, often assisted by monocular or RGB-D geometric priors~\cite{monoscene,ndcscene,ISO,fbocc,lss,piccinelli2025unidepthv2,shao2024nddepth}, to more recent transformer-augmented volumetric models and sparse Gaussian or point-based representations~\cite{gaussianformer,embodiedocc,embodiedocc++,roboocc}.
The former typically rely on 3D CNNs or volumetric heads for decoding, while the latter progressively refine geometric and semantic representations to improve efficiency and prediction fidelity.
However, indoor environments differ substantially from structured road scenes in layout complexity, scale and viewpoint variation, severe occlusion, and the long-tailed distribution of fine-grained semantic categories.
Consequently, designs tailored to autonomous-driving scenarios may transfer suboptimally to indoor environments, motivating further investigation into feature lifting, spatial representations, multi-view geometry, and supervision paradigms~\cite{ISO,embodiedocc,roboocc,xie2025benchmarking,sarlin2023pixel}.

\subsection{Open-Vocabulary Occupancy Prediction}

Open-vocabulary occupancy prediction has recently gained increasing attention in autonomous driving as a means of reducing the cost and difficulty of dense 3D occupancy annotation.
Existing methods typically adopt self-supervised or weakly supervised settings to learn voxel-level geometry and semantics without full 3D supervision.
For semantic learning, most approaches first obtain open-vocabulary cues from vision--language models and 2D regions from promptable or open-set segmentation models~\cite{groundedsam,groundingdino,liu2025explicit}.
These semantic signals are then transferred to 3D in different ways: by projecting image pixels onto point clouds or associating pixels and points with voxels~\cite{occnerf,li2025ago,yu2025language}, by mapping image features to voxels through differentiable projection~\cite{veon,pop3d}, or by directly supervising rendered 3D semantic fields in the image plane via volume rendering~\cite{boeder2024langocc}.
Feature alignment is commonly performed between learned 3D representations and language-aligned image or text embeddings using cosine similarity~\cite{groundedsam,groundingdino}.
For geometry learning, existing approaches mainly follow two paradigms: LiDAR-derived pseudo labels aggregated into voxels for supervision~\cite{veon,gao2025loc,pop3d,li2025ago}, and self-supervised multi-view rendering that learns occupancy by enforcing photometric, feature, or geometric consistency across views without explicit 3D ground-truth occupancy~\cite{selfocc,occnerf,sarlin2023pixel}.
Compared with outdoor driving scenarios, open-vocabulary occupancy prediction in indoor environments remains relatively underexplored.
OVO~\cite{ovo} provides an early attempt on a small-scale dataset by distilling open-vocabulary semantics into a 3D occupancy network through pixel-wise alignment between projected voxel features and 2D open-vocabulary segmentation.
OpenOcc~\cite{OpenOcc} further leverages neural radiance fields to jointly model 3D scene reconstruction and semantic understanding from RGB-D inputs, while OG~\cite{dong2023og} integrates Grounded-SAM~\cite{groundedsam} with affinity fields to distinguish grounded object instances.
Despite these efforts, scalable open-vocabulary occupancy prediction for large-scale indoor scenes remains largely underexplored.

\section{Method}\label{sec:method}

\subsection{Problem Setting}\label{sec:method_overview}
\begin{figure*}[h]
  \centering
  \includegraphics[width=0.98\textwidth]{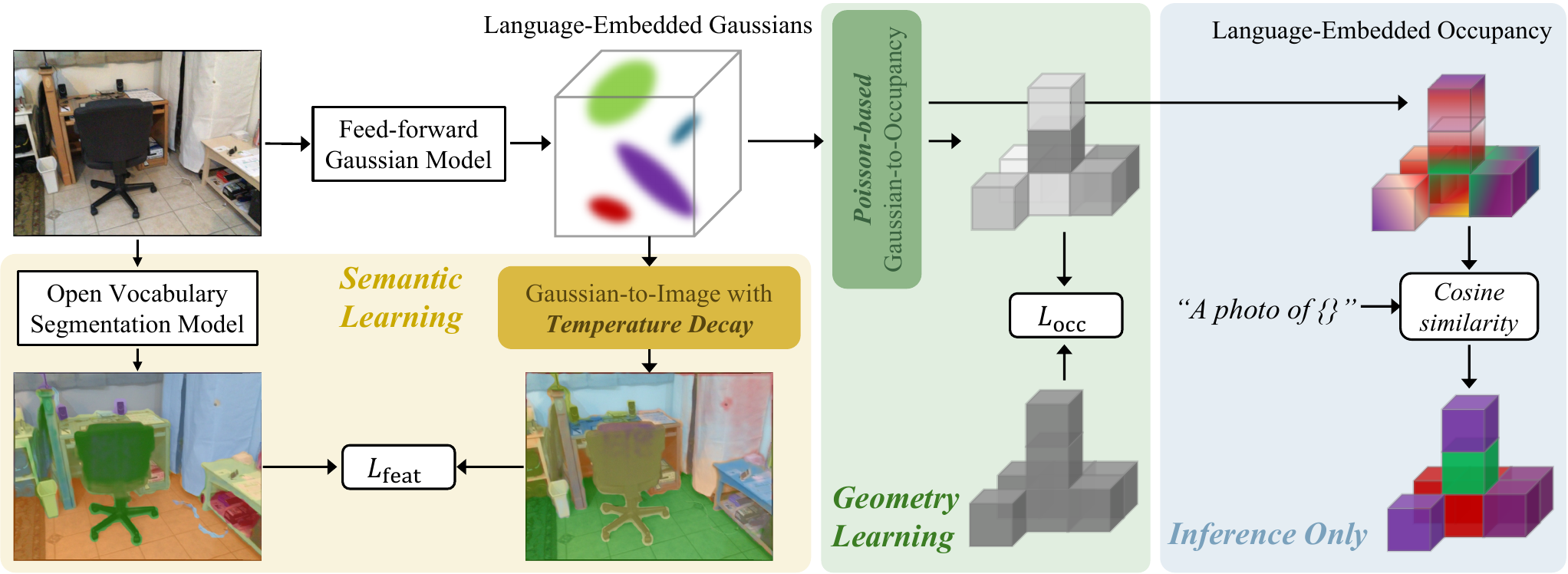}
  \vspace{-1mm}
  \caption{\textbf{Framework Overview.} From a monocular image, a feed-forward Gaussian model produces \emph{Language-Embedded Gaussians}. Training proceeds along two paths: \emph{Semantic learning}, we differentiably render Gaussian features to the image with \emph{Progressive Temperature Decay} and align them to a training-free open-vocabulary segmenter via a cosine objective $L_{\text{feat}}$; \emph{Geometry learning}, we convert Gaussians to occupancy using an opacity-aware Poisson-based Gaussian-to-Occupancy operator and supervise binary occupancy with $L_{\text{occ}}$. At inference, the Language-Embedded Occupancy supports text-driven queries by computing cosine similarity between voxel embeddings and prompt embeddings, yielding open-vocabulary semantic occupancy without dense voxel-level semantic labels during training.}
  \label{fig:framework}
\end{figure*}

Given an input image $\mathbf{I}\in\mathbb{R}^{H\times W\times 3}$, the goal of occupancy estimation is to predict a semantic occupancy field $\mathbf{O}\in\mathbb{R}^{X\times Y\times Z\times C}$, where $(X,Y,Z)$ specify the resolution of the local 3D volume and $C$ is the number of semantic categories. We decompose the problem into two complementary components: (i) \emph{geometry learning}, which determines whether each voxel is occupied, and (ii) \emph{semantic learning}, which assigns a category to occupied voxels.

To enable open-vocabulary occupancy, we employ \emph{Language-Embedded Gaussians} (LE-Gaussians) as a unified 3D intermediate representation coupling geometry and language-aligned semantics. 
Each LE-Gaussian is parameterized as:
\begin{equation}\label{eq:gauss_3d}
\mathcal{G}_i=\big(\boldsymbol{\mu}_i,\boldsymbol{\Sigma}_i,\alpha_i,\mathbf{f}_i\big),
\end{equation}
where $\boldsymbol{\mu}_i\in\mathbb{R}^3$ and $\boldsymbol{\Sigma}_i\in\mathbb{R}^{3\times 3}$ are the Gaussian center and covariance, $\alpha_i\in[0,1]$ denotes opacity, and $\mathbf{f}_i\in\mathbb{R}^d$ is a language-aligned embedding with $d$ channels.
LE-Gaussians thus explicitly bind \emph{geometry} (via $\boldsymbol{\mu}_i,\boldsymbol{\Sigma}_i,\alpha_i$) and \emph{open-vocabulary semantics} (via $\mathbf{f}_i$) within a single set of primitives, supporting both occupancy prediction and semantic reasoning.

\subsection{Framework Overview}\label{sec:overview}
As illustrated in~\Cref{fig:framework}, given an input image $\mathbf{I}$, a feed-forward Gaussian predictor outputs a set of LE-Gaussians $\mathbf{G}=\{\mathcal{G}_i\}_{i=1}^{N}$. The same set $\mathbf{G}$ is then reused for two purposes: 1) \emph{occupancy prediction} (\Cref{sec:g2o}), where our Poisson-based Gaussian-to-Occupancy (G2O) operator infers 3D occupancy from geometric parameters $(\boldsymbol{\mu}_i,\boldsymbol{\Sigma}_i,\alpha_i)$; and
2) \emph{language-aligned semantic prediction}, where per-Gaussian embeddings $\mathbf{f}_i$ are used to infer the semantic label of each occupied voxel. Supervision is imposed by rendering these embeddings to the image plane and aligning the rendered features with those from open-vocabulary segmentation models; the alignment is further improved by our progressive temperature decay (\Cref{sec:gs_temp}).
Training jointly minimizes a supervised binary-occupancy loss with ground-truth and a rendered-feature alignment loss that requires no 2D human annotations; see \Cref{sec:loss} for details.

\subsection{Poisson-based Gaussian-to-Occupancy}\label{sec:g2o}

3D Gaussians offer an efficient and expressive representation to capture fine-grained indoor geometry
and have been widely adopted for scene reconstruction. In this section, we first revisit Gaussian splatting, analyze why existing G2O formulations become problematic under binary-only occupancy supervision, and then introduce our Poisson-based G2O reformulation that addresses the issue.

\noindent\textbf{Gaussian Splatting.}
We follow the front-to-back $\alpha$-blending strategy of Gaussian Splatting~\cite{gaussiansplatting} to render features from LE-Gaussians.
For the $i$-th Gaussian $\mathcal{G}_i$ with feature $\mathbf{f}_i$, opacity $\alpha_i$, and screen-space mean/covariance $(\boldsymbol{\mu}'_i,\boldsymbol{\Sigma}'_i)$, its 2D kernel at image location $\mathbf{x}'$ is
\begin{equation}
p_i(\mathbf{x}') \;=\; \exp\!\Big(-\tfrac{1}{2}\,(\mathbf{x}'-\boldsymbol{\mu}'_i)^{\!\top}\,{\boldsymbol{\Sigma}'_i}^{-1}\,(\mathbf{x}'-\boldsymbol{\mu}'_i)\Big),
\end{equation}
where primes denote screen-space quantities corresponding to the 3D parameters in \Cref{eq:gauss_3d}.
We define the \emph{effective opacity} at $\mathbf{x}'$ as
\begin{equation}
\tilde{\alpha}_i(\mathbf{x}') \;=\; \alpha_i\, p_i(\mathbf{x}').
\end{equation}
and render the pixel feature by $\alpha$-blending:
\begin{equation}
\mathbf{F}(\mathbf{x}')
\;=\;
\sum_{i=1}^{N}
\Bigg(\prod_{j=1}^{i-1}\!\big(1-\tilde{\alpha}_j(\mathbf{x}')\big)\Bigg)\,
\tilde{\alpha}_i(\mathbf{x}')\,\mathbf{f}_i.
\label{eq:a_blend}
\end{equation}

\noindent\textbf{Limitations of GaussianFormer2.}
To infer occupancy from Gaussians, GaussianFormer2~\cite{gaussianformer2} introduces a \emph{probabilistic Gaussian superposition} that treats each primitive $\mathcal{G}_i$ as a local occupancy distribution and aggregates them multiplicatively. Concretely, the contribution of $\mathcal{G}_i$ at a 3D location $\mathbf{x}$ is
\begin{equation}
p_i(\mathbf{x}) \;=\; \exp\!\Big(-\tfrac{1}{2}\,(\mathbf{x}-\boldsymbol{\mu}_i)^{\!\top}\,\boldsymbol{\Sigma}_i^{-1}\,(\mathbf{x}-\boldsymbol{\mu}_i)\Big),
\label{eq:single-prob}
\end{equation}
and, assuming independence across Gaussians, the aggregated occupancy becomes
\begin{equation}
p(\mathbf{x}) \;=\; 1 - \prod_{i=1}^{N}\!\big(1 - p_i(\mathbf{x})\big).
\label{eq:multi-prob}
\end{equation}
While efficient and simple, this formulation omits opacity $\alpha_i$ in the geometry branch and relies solely on the spatial kernels.
As a result, the occupancy inference is decoupled from the opacity used in image rendering, creating a mismatch between 2D rendering and 3D voxel aggregation.

\noindent\textbf{Why is this problematic?}
In GaussianFormer2, geometry is aggregated without opacity (Eq.~\ref{eq:multi-prob}), but this does not cause issues because: 1) their semantic learning explicitly models opacity, and 2) it does not rely on 2D rendering during training, so no inconsistency arises.
By contrast, our semantics are learned by \emph{rendering} LE-Gaussian (Eq.~\ref{eq:a_blend}) to images and aligning them to 2D open-vocabulary segmentations features, which is inherently opacity-aware.
Using an opacity-agnostic voxel aggregation for geometry while supervising semantics via opacity-aware rendering introduces conflicting signals and unstable optimization.

\noindent\textbf{A Bernoulli approach.}
A direct fix is to replace the spatial kernel $p_i(\mathbf{x})$ with the \emph{effective opacity} $\tilde{\alpha}_i(\mathbf{x})=\alpha_i\,p_i(\mathbf{x})$ and reuse the complementary-probability rule:
\begin{equation}
p(\mathbf{x})
= 1-\prod_{i=1}^{N}\!\big(1-\tilde{\alpha}_i(\mathbf{x})\big)
= 1-\prod_{i=1}^{N}\!\big(1-\alpha_i\,p_i(\mathbf{x})\big),
\label{eq:naive_opacity}
\end{equation}
which can be viewed as independent Bernoulli ''hits`` from overlapping Gaussians at $\mathbf{x}$.
A similar opacity-based idea appears in Gaussian Opacity Fields~\cite{Yu2024GOF} and in concurrent fully supervised closed-vocabulary method~\cite{qian2025splatsscdecoupleddepthguidedgaussian,park2025s2gostreamingsparsegaussian}.

However, we observe that the Bernoulli aggregation drives the learned opacities to \emph{small} values, which enlarges the gap between rendered features and per-Gaussian embeddings (see \Cref{sec:gs_temp}).
This behavior likely arises because when multiple Gaussians overlap within a voxel, the Bernoulli union $1-\prod_i(1-\alpha_i p_i)$ saturates rapidly toward~1; and for a fixed total mass $\sum_i \alpha_i p_i$, the union is maximized when each of the overlapping contributions are similar, making it difficult to learn differentiated opacities.

\paragraph{A Poisson approach.}
We adopt a Poisson formulation that treats each Gaussian’s local contribution as a nonnegative \emph{event intensity} (mean count):
\begin{equation}
h_i(\mathbf{x}) \;\triangleq\; \alpha_i\,p_i(\mathbf{x}) \;\ge 0,
\qquad
z(\mathbf{x}) \;=\; \sum_{i=1}^{N} h_i(\mathbf{x}) ,
\label{eq:hgs_hz}
\end{equation}
where $h_i$ is the expected contribution of Gaussian $\mathcal{G}_i$ over a small voxel segment, and $z(\mathbf{x})$ is the \emph{mean measure} (cumulative expected count) from all Gaussians at $\mathbf{x}$.
Modeling occupancy as the probability that a non-homogeneous Poisson process (NHPP) has produced at least one event at location $\mathbf{x}$ gives a simple closed form.
Let $z(\mathbf{x})$ denote the accumulated mean measure at $\mathbf{x}$.
A classical result for Poisson processes states that the probability of observing no event is
$\Pr[\text{no event at }\mathbf{x}] = \exp\!\big(-z(\mathbf{x})\big)$~\cite{ross1996stochastic,kingman-poisson-processes}.
Interpreting “at least one event” as the space being occupied, the occupancy becomes a single exponential of the total mean measure:
\begin{equation}
p(\mathbf{x})
= 1-\exp\!\big(-z(\mathbf{x})\big)
= 1-\exp\!\Big(-\sum_{i=1}^{N}\alpha_i\,p_i(\mathbf{x})\Big).
\label{eq:hgs_occ}
\end{equation}

\subsection{Progressive Temperature Decay}\label{sec:gs_temp}

As shown in the semantic learning part of~\Cref{fig:framework}, we adopt Gaussian Splatting~\cite{gaussiansplatting} to render LE-Gaussians into images. 
However, the $\alpha$-blending in~\Cref{eq:a_blend} produces a \emph{weighted sum} of per-ray contributions, so each pixel feature becomes a mixture of multiple Gaussian embeddings along the ray. Consequently, the supervision signal encourages the rendered mixture, rather than each individual Gaussian, to be language-aligned, creating a mismatch that later hampers G2O-based splatting. This ambiguity is further exacerbated in indoor scenes, where objects are numerous, fine-grained, and heavily overlapping in projection.
Dr.\ Splat~\cite{drsplat25} mitigates this via hard \emph{feature registration}: along a pixel ray it assigns the pixel’s CLIP embedding to only the Top-$k$ dominant Gaussians, yielding language-aligned vectors that are directly retrievable in 3D. While effective, this discrete selection causes sparse gradients, since supervision backpropagates only through the chosen Top-$k$ primitives.

\begin{figure}[h!]
  \centering
  \includegraphics[width=0.4\textwidth]{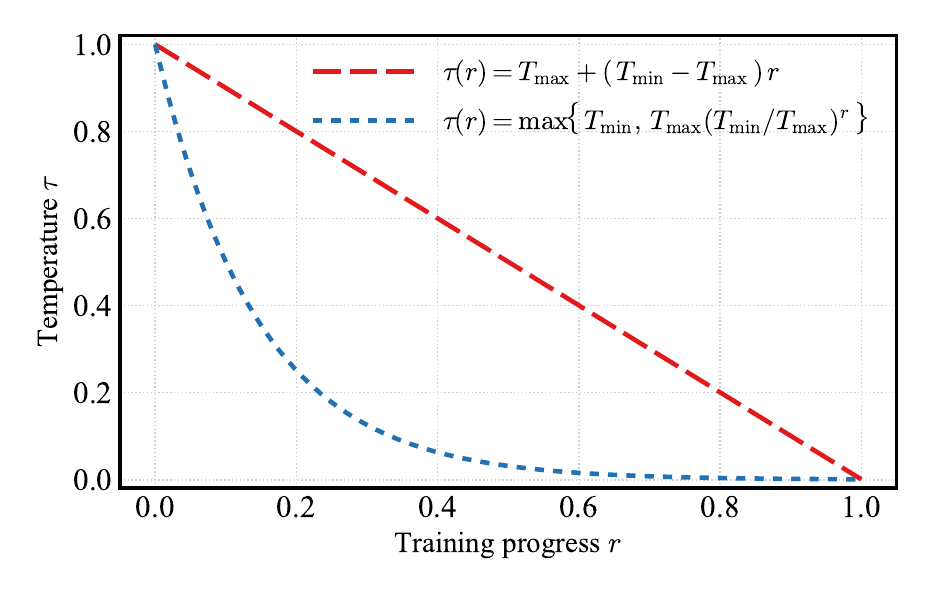}
  \vspace{-2mm}
 \caption{\textbf{Comparison of temperature schedules.} Linear decay decreases $\tau$ uniformly, whereas our exponential schedule rapidly approaches $T_{\min}$, allocating more iterations for the model to adapt to the low-temperature regime.}
  \label{fig:temp_decay}
\end{figure}

Instead of a hard Top-$k$, we reduce mixture ambiguity via a continuous and training-stable \emph{opacity sharpening}.
Let $\alpha_i=\sigma(\alpha_i^{\text{logit}})$, where $\sigma$ is the sigmoid function and $\alpha_i^{\text{logit}}$ is the predicted logit. We adopt a \emph{tempered} sigmoid
\begin{equation}
\alpha_i \;=\; \sigma\!\left(\frac{\alpha_i^{\text{logit}}}{\tau}\right),
\end{equation}
where smaller (near 0) temperature $\tau$ values sharpen opacities
towards $\{0,1\}$, reducing feature mixing along the ray while maintaining gradient flow to all contributors (rather than a discrete subset).
In practice, we employ a \emph{progressive temperature decay} schedule during training, annealing $\tau$ from $T_{\max}$ to $T_{\min}$ (default $T_{\max}{=}1$, $T_{\min}{=}10^{-3}$). Let $r\!\in[0,1]$ denote the normalized training progress, 0 means beginning, 1 means end. We use an \emph{exponential} schedule
\[
\tau(r)\;=\;\max\!\Big\{T_{\min},\;T_{\max}\,(T_{\min}/T_{\max})^r\Big\},
\]
which starts with smooth mixtures for stable optimization and gradually sharpens opacities for discriminative alignment. 
Compared with linear decay, our exponential schedule naturally allocates more iterations at low temperatures (see \Cref{fig:temp_decay}), empirically yielding \emph{sharper} per-Gaussian opacities and stronger language alignment while preserving end-to-end differentiability throughout. 
With the proposed temperature-decay schedule, as $\tau$ decreases, mixture blur is gradually reduced and supervision is increasingly transferred to individual Gaussians.

\subsection{Instance-Level Semantic Consistency}
\label{sec:sam_instance_consistency}

\begin{figure}[h]
  \centering
  \includegraphics[width=\columnwidth]{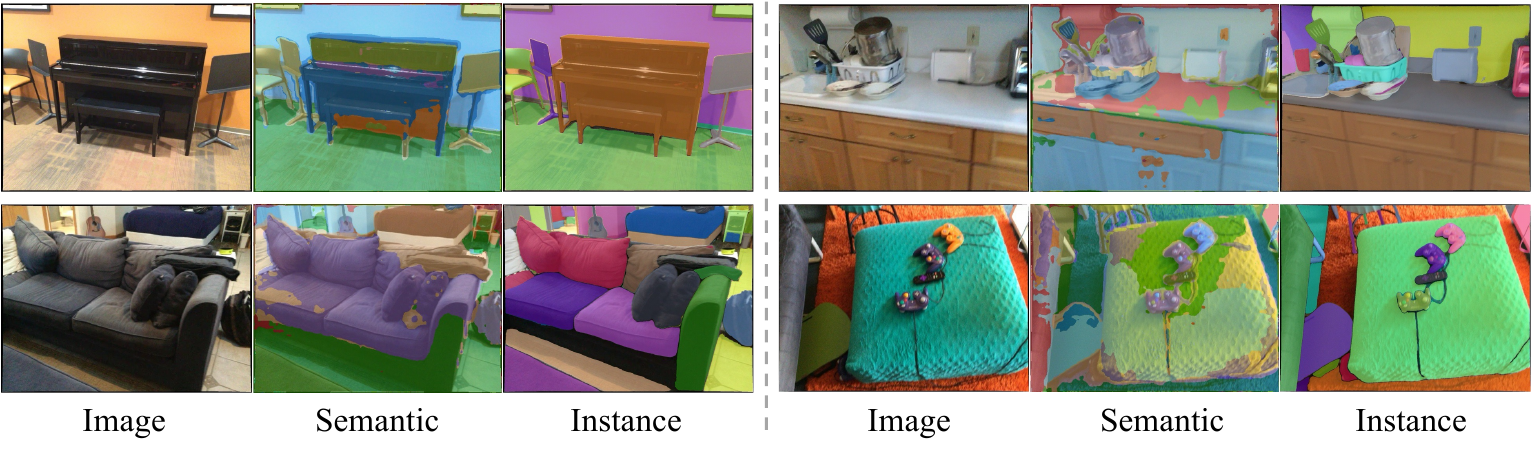}
  \vspace{-3mm}
  \caption{
  \textbf{Instance-level semantic consistency.}
  Given an input image (left), a frozen open-vocabulary teacher (e.g., Trident~\cite{trident}) provides semantic cues in image space (middle), which can be spatially fragmented and incomplete for the same object. In contrast, SAM produces category-agnostic instance masks (right) that more coherently capture complete object extents.
  We aggregate teacher features within each reliable SAM mask to form an instance-level semantic prototype, and encourage rendered LE-Gaussian features inside the same instance to align with this shared prototype.
  }
  \label{fig:inst_consistency}
\end{figure}

The pixel-wise feature alignment transfers open-vocabulary semantics to LE-Gaussians, but treats each rendered pixel independently. As a result, semantic predictions can be spatially fragmented: different regions of the same object may receive inconsistent semantic embeddings, or even be divided into multiple disconnected semantic regions. As illustrated in~\Cref{fig:inst_consistency}, the open-vocabulary teacher provides rich semantic cues, but its predictions may only cover fragmented parts of an object. In contrast, SAM produces category-agnostic instance masks that more coherently capture the complete spatial extent of individual objects.
We therefore complement the original pixel-level supervision with an \emph{instance-level semantic consistency} objective, using SAM masks to aggregate fragmented semantic evidence and encourage features within the same instance to share a coherent semantic representation.
Let $\mathcal{M}_v$ denote the set of SAM masks in view $v$.
For each mask $M\in\mathcal{M}_v$, we retain only pixels with sufficiently high rendered opacity:
\begin{equation}
\Omega_M
=
\left\{
\mathbf{x}\in M
\;\middle|\;
\alpha_v(\mathbf{x}) > \alpha_{\mathrm{inst}}
\right\},
\end{equation}
where $\alpha_v(\mathbf{x})$ is the rendered opacity at pixel $\mathbf{x}$.
This removes pixels that are weakly explained by the rendered Gaussians. Since SAM may produce small or fragmented masks with only a few valid rendered pixels, their aggregated semantic prototypes can be unreliable.
We therefore require each mask to contain at least $\tau_{\mathrm{inst}}$ valid pixels, and discard masks with $|\Omega_M| < \tau_{\mathrm{inst}}$.
Let $\hat{\mathbf{F}}^{\,\mathrm{T}}_v(\mathbf{x})$ denote the $\ell_2$-normalized feature from the frozen open-vocabulary teacher.
For each retained instance mask, we aggregate the teacher features within the mask into an instance-level semantic prototype:
\begin{equation}
\mathbf{c}_M
=
\frac{
\sum_{\mathbf{x}\in\Omega_M}
\hat{\mathbf{F}}^{\,\mathrm{T}}_v(\mathbf{x})
}{
\left\|
\sum_{\mathbf{x}\in\Omega_M}
\hat{\mathbf{F}}^{\,\mathrm{T}}_v(\mathbf{x})
\right\|_2
}.
\label{eq:instance_prototype}
\end{equation}
This prototype consolidates potentially fragmented local semantic cues into a shared representation for the entire object instance.
We then encourage every rendered feature within the instance to align with the shared prototype:
\begin{equation}
L_{\mathrm{inst}}
=
\frac{1}{|\mathcal{M}_v|}
\sum_{M\in\mathcal{M}_v}
\frac{1}{|\Omega_M|}
\sum_{\mathbf{x}\in\Omega_M}
\left(
1-
\hat{\mathbf{F}}_v(\mathbf{x})^{\top}\mathbf{c}_M
\right),
\label{eq:sam_inst_loss}
\end{equation}
where $\hat{\mathbf{F}}_v(\mathbf{x})$ is the normalized feature rendered from LE-Gaussians.
Unlike the original pixel-wise feature loss, which preserves fine-grained local semantic supervision, $L_{\mathrm{inst}}$ leverages the more complete object support provided by SAM to impose a complementary object-level constraint, reducing semantic fragmentation and improving feature consistency within the same instance.

\subsection{Covisibility-Aware Support-View Selection}
\label{sec:covis_frame_selection}

\begin{figure}[h!]
  \centering
  \includegraphics[width=0.5\textwidth]{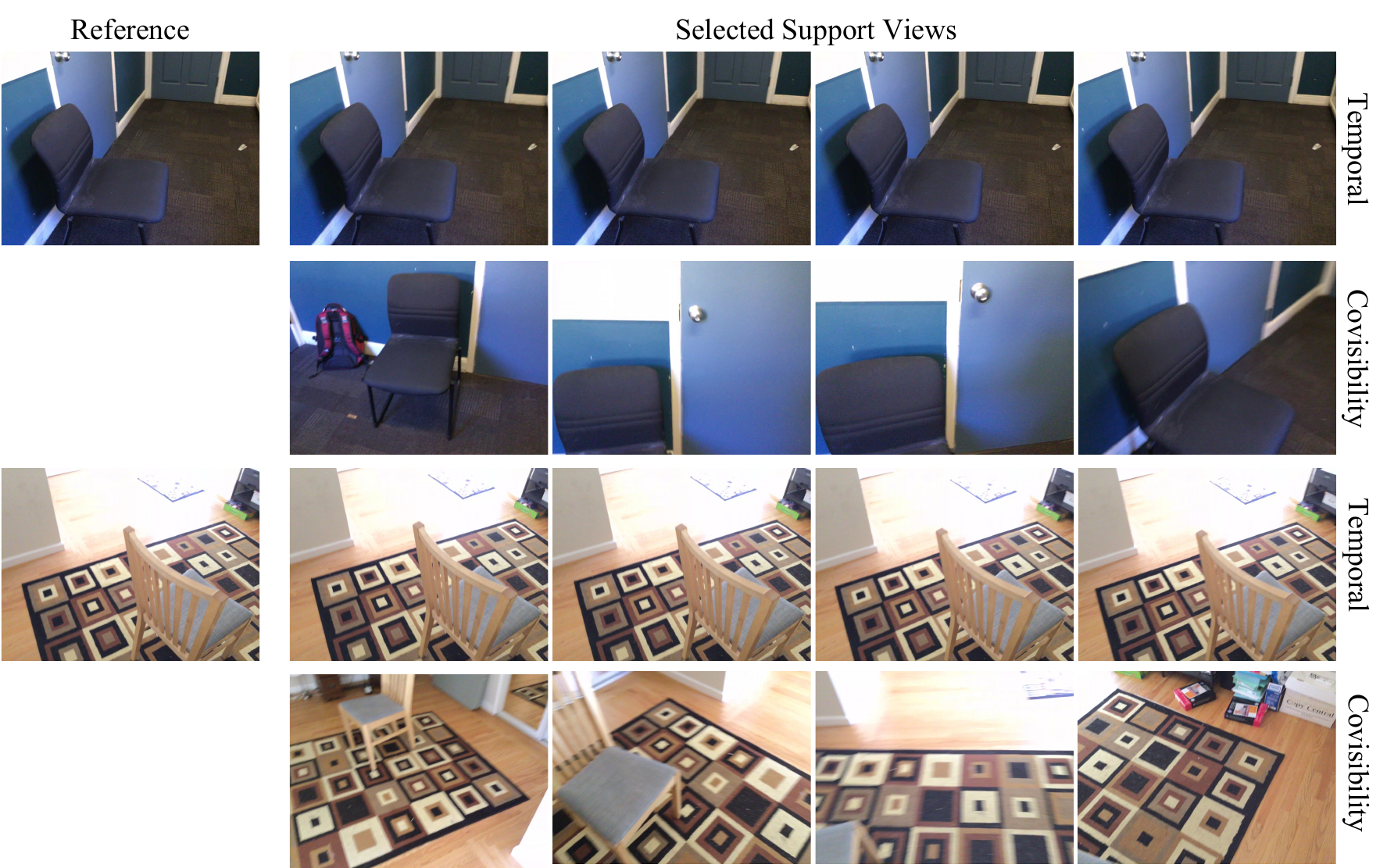}
  \vspace{-3mm}
  \caption{
\textbf{Covisibility-aware support-view selection.}
Each example shows a reference view (left) and four support views. Fixed temporal selection often produces highly redundant observations (top). Our method filters views with either insufficient overlap or near-duplicate content according to geometric covisibility, and selects pose-diverse views from the remaining candidates (bottom), providing complementary observations while preserving shared scene geometry.}
  \label{fig:covis_sample}
\end{figure}

Multi-view feature alignment benefits from support views that preserve sufficient overlap with the reference view while providing complementary observations. A straightforward strategy is to select nearby frames using a fixed temporal interval~\cite{legoocc}.
However, as illustrated in~\Cref{fig:covis_sample}, a small temporal interval often yields highly similar and redundant views, whereas increasing the interval may lead to insufficient scene overlap. Moreover, a fixed frame gap is difficult to generalize across videos with different frame rates and sampling frequencies.
We therefore select support views according to \emph{geometric covisibility}.
For each frame $i$, valid depth pixels are back-projected into 3D and discretized into a set of visible voxels $\mathcal{V}_i$.
We define the covisibility between a reference view $t$ and a candidate view $s$ as
\begin{equation}
c_{t,s}
=
\frac{
|\mathcal{V}_t\cap\mathcal{V}_s|
}{
\min\!\left(|\mathcal{V}_t|,|\mathcal{V}_s|\right)
}.
\label{eq:covis_score}
\end{equation}
This score measures the proportion of shared visible geometry between the two views.
We retain candidates satisfying
$c_{\min}\le c_{t,s}\le c_{\max}$:
the lower bound removes views with insufficient overlap, while the upper bound suppresses near-duplicate observations.
From the remaining candidates, we greedily select pose-diverse support views to encourage complementary cross-view supervision.
As shown in~\Cref{fig:covis_sample}, the resulting views preserve sufficient shared geometry with the reference frame while providing substantially more diverse viewpoints than fixed-interval selection.
In our default setting, we use $c_{\min}{=}0.2$ and $c_{\max}{=}0.6$.

\subsection{Losses}\label{sec:loss}

We optimize the network with geometry and language-aligned semantic supervision, with different geometry objectives for indoor and outdoor settings.
For indoor scenes, following~\cite{embodiedocc}, we supervise geometry with ground-truth depth and binary voxel occupancy labels using focal loss $L_{\mathrm{focal}}$, Lovász--Softmax loss $L_{\mathrm{lov}}$, and scene-class affinity loss $L_{\mathrm{scal}}$.
For outdoor scenes, we follow the \textbf{annotation-free} setting and do not use any occupancy annotations; geometry is supervised by multi-view rendering and a Huber depth loss $L_{\mathrm{depth}}$ using pseudo-depth generated by the frozen Metric3D~\cite{yin2023metric3d} model.
For semantic learning, $L_{\mathrm{feat}}$ aligns features rendered from LE-Gaussians with those from the frozen open-vocabulary teacher, including the support views selected by our covisibility-aware strategy (\Cref{sec:covis_frame_selection}).
We further apply the SAM-guided instance consistency loss $L_{\mathrm{inst}}$ (\Cref{sec:sam_instance_consistency}) to encourage object-level semantic coherence.
Accordingly, the indoor objective is
\begin{equation}
\label{eq:loss_indoor}
\begin{aligned}
L_{\mathrm{indoor}}
\;&=\;
\lambda_{\mathrm{focal}}L_{\mathrm{focal}}
+
\lambda_{\mathrm{lov}}L_{\mathrm{lov}}
+
\lambda_{\mathrm{scal}}L_{\mathrm{scal}} \\
&\quad+
\lambda_{\mathrm{depth}}L_{\mathrm{depth}}
+
\lambda_{\mathrm{feat}}L_{\mathrm{feat}}
+
\lambda_{\mathrm{inst}}L_{\mathrm{inst}},
\end{aligned}
\end{equation}
while the outdoor objective removes the occupancy-supervision terms:
\begin{equation}
\label{eq:loss_outdoor}
L_{\mathrm{outdoor}}
=
\lambda_{\mathrm{depth}}L_{\mathrm{depth}}
+
\lambda_{\mathrm{feat}}L_{\mathrm{feat}}
+
\lambda_{\mathrm{inst}}L_{\mathrm{inst}}.
\end{equation}

\section{Experiments}\label{sec:exp}

\begin{table*}[t]
    \caption{\textbf{Monocular results on Occ-ScanNet.}
    $^{\dagger}$We re-implement POP-3D~\cite{pop3d} and LOcc~\cite{yu2025language} for the indoor monocular setting, and \textbf{supervise geometry with ground-truth binary occupancy} instead of the authors’ LiDAR-derived pseudo labels. We also adopt DepthAnything-v2~\cite{depthanythingv2} as the backbone and use the same input resolution and training recipe as ours for fair comparison. 
    For POP-3D, we project voxel centers to the image plane and sample language-aligned features at those locations to form a 3D grid of text-aligned embeddings for open-vocabulary reasoning.
    For LOcc, we prompt Qwen2.5-VL-7B~\cite{Qwen2.5-VL} to extract object names and use Trident~\cite{trident} to obtain training-free open-vocabulary segmentations.
    }
    \vspace{-2mm}
    \setlength{\tabcolsep}{0.008\textwidth}
    \captionsetup{font=scriptsize}
    \begin{center}
    \resizebox{0.95\linewidth}{!}{
    \begin{tabular}{l|c|c c c c c c c c c c c|c}
        \toprule
        Method
        & {IoU}
        & \rotatebox{90}{\parbox{1.5cm}{\textcolor{ceiling}{$\blacksquare$} ceiling}} 
        & \rotatebox{90}{\textcolor{floor}{$\blacksquare$} floor}
        & \rotatebox{90}{\textcolor{wall}{$\blacksquare$} wall} 
        & \rotatebox{90}{\textcolor{window}{$\blacksquare$} window} 
        & \rotatebox{90}{\textcolor{chair}{$\blacksquare$} chair} 
        & \rotatebox{90}{\textcolor{bed}{$\blacksquare$} bed} 
        & \rotatebox{90}{\textcolor{sofa}{$\blacksquare$} sofa} 
        & \rotatebox{90}{\textcolor{table}{$\blacksquare$} table} 
        & \rotatebox{90}{\textcolor{tvs}{$\blacksquare$} tvs} 
        & \rotatebox{90}{\textcolor{furniture}{$\blacksquare$} furniture} 
        & \rotatebox{90}{\textcolor{objects}{$\blacksquare$} objects} 
        & mIoU\\
        \midrule
        \rowcolor{Sbg}\multicolumn{14}{l}{\textit{Closed-vocabulary (full annotation)}}\\
        \arrayrulecolor{lightgray}\midrule\arrayrulecolor{black}
        \rowcolor{Sbg} TPVFormer~\cite{Triformer} & 33.39 & 6.96 & 32.97 & 14.41 & 9.10 & 24.01 & 41.49 & 45.44 & 28.61 & 10.66 & 35.37 & 25.31 & 24.94 \\
        \rowcolor{Sbg} GaussianFormer~\cite{gaussianformer} & 40.91 & 20.70 & 42.00 & 23.40 & 17.40 & 27.0 & 44.30 & 44.80 & 32.70 & 15.30 & 36.70 & 25.00 & 29.93 \\
        \rowcolor{Sbg} MonoScene~\cite{monoscene} & 41.60 & 15.17 & 44.71 & 22.41 & 12.55 & 26.11 & 27.03 & 35.91 & 28.32 & 6.57 & 32.16 & 19.84 & 24.62 \\
        \rowcolor{Sbg} ISO~\cite{ISO} & 42.16 & 19.88 & 41.88 & 22.37 & 16.98 & 29.09 & 42.43 & 42.00 & 29.60 & 10.62 & 36.36 & 24.61 & 28.71 \\
        \rowcolor{Sbg} Surroundocc~\cite{surroundocc} & 42.52 & 18.90 & 49.30 & 24.80 & 18.00 & 26.80 & 42.00 & 44.10 & 32.90 & 18.60 & 36.80 & 26.90 & 30.83 \\
        \rowcolor{Sbg} EmbodiedOcc~\cite{embodiedocc} & 53.55 & 39.60 & 50.40 & 41.40 & 31.70 & 40.90 & 55.00 & 61.40 & 44.00 & 36.10 & 53.90 & 42.20 & 45.15 \\
        \rowcolor{Sbg} EmbodiedOcc++~\cite{embodiedocc++} & 54.90 & 36.40 & 53.10 & 41.80 & 34.40 & 42.90 & 57.30 & 64.10 & 45.20 & 34.80 & 54.20 & 44.10 & 46.20 \\
        \rowcolor{Sbg} RoboOcc~\cite{roboocc} & 56.48 & 45.36 & 53.49 & 44.35 & 34.81 & 43.38 & 56.93 & 63.35 & 46.35 & 36.12 & 55.48 & 44.78 & 47.76 \\
        \midrule
        \rowcolor{Ubg}\multicolumn{14}{l}{\textit{Open-vocabulary (geometry-only annotation)}}\\
        \arrayrulecolor{lightgray}\midrule\arrayrulecolor{black}
        \rowcolor{Ubg} POP-3D$^{\dagger}$~\cite{pop3d} & 35.32 & 0.00 & 23.89 & 16.59 & 0.01 & 4.62 & 8.65 & 2.01 & 7.61 & 0.00 & 2.14 & 0.08 & 5.96  \\
        \rowcolor{Ubg} LOcc$^{\dagger}$~\cite{yu2025language} & 36.70 & 0.00 & 24.40 & 17.37 & 1.83 & 9.31 & 10.73 & 23.42 & 11.29 & 0.01 & 3.15 & 0.19 & 9.25 \\
        \rowcolor{UbgStrong} \modelname{} (Ours) & 59.50 & 17.81 & 41.65 & 28.52 & 18.91 & 19.32 & 33.05 & 34.92 & 19.16 & 5.36 & 5.88 & 6.94 & 21.05 \\
        \rowcolor{UbgStrong} \newmodelname{} (Ours) & 61.73 & 31.13 & 42.80 & 29.53 & 21.97 & 23.87 & 36.17 & 39.71 & 22.11 & 8.51 & 3.65 & 7.00 & 24.22 \\
        \bottomrule
    \end{tabular}
    }
    \end{center}
    \vspace{-3mm}
    \label{tab:main_mono}
\end{table*}

\begin{table*}[h]
    \setlength{\belowcaptionskip}{8pt}
    \setlength{\abovecaptionskip}{4pt}
    \caption{
    \textbf{Quantitative performance of 3D occupancy methods on the Occ3D-nuScenes dataset.}
    We compare methods that do not rely on LiDAR supervision.
    \emph{Frames} denotes the number of temporal frames required at inference, and methods requiring more than one frame are shown in gray.
    IoU scores for ``others'' and ``other flat'' classes are consistently zero and thus omitted.
    ``Cons. veh.'' denotes construction vehicles, and ``drive. surf.'' denotes drivable surfaces.
    }
    \label{tab:main_nusc}
    \centering
    \resizebox{\textwidth}{!}{
    \begin{tabular}{
        l | c | c |
        r r r r r r r r r r r r r r r
    }
        \toprule[1.2pt]
        Method
        & Frames
        & mIoU
        & \clsname{barrier}{barrier}
        & \clsname{bicycle}{bicycle}
        & \clsname{bus}{bus}
        & \clsname{car}{car}
        & \clsname{cons. veh.}{construct}
        & \clsname{motorcycle}{motor}
        & \clsname{pedestrian}{pedestrian}
        & \clsname{traffic cone}{traffic}
        & \clsname{trailer}{trailer}
        & \clsname{truck}{truck}
        & \clsname{drive. surf.}{driveable}
        & \clsname{sidewalk}{sidewalk}
        & \clsname{terrain}{terrain}
        & \clsname{manmade}{manmade}
        & \clsname{vegetation}{vegetation}
        \\
        \midrule

        SelfOcc~\cite{selfocc}
        & 1
        & 10.54
        & 0.15
        & 0.66
        & 5.46
        & 12.54
        & 0.00
        & 0.80
        & 2.10
        & 0.00
        & 0.00
        & 8.25
        & 55.49
        & 26.30
        & 26.54
        & 14.22
        & 5.60
        \\

        OccNeRF~\cite{occnerf}
        & 1
        & 10.81
        & 0.83
        & 0.82
        & 5.13
        & 12.49
        & 3.50
        & 0.23
        & 3.10
        & 1.84
        & 0.52
        & 3.90
        & 52.62
        & 20.81
        & 24.75
        & 18.45
        & 13.19
        \\
        GaussianOcc~\cite{gan2025gaussianocc}
        & 1
        & 11.26
        & 1.79
        & 5.82
        & 14.58
        & 13.55
        & 1.30
        & 2.82
        & 7.95
        & 9.76
        & 0.56
        & 9.61
        & 44.59
        & 20.10
        & 17.58
        & 8.61
        & 10.29
        \\
        \graycell{LangOcc~\cite{boeder2024langocc}}
        & \graycell{2}
        & \graycell{12.04}
        & \graycell{2.70}
        & \graycell{7.20}
        & \graycell{5.80}
        & \graycell{13.90}
        & \graycell{0.50}
        & \graycell{10.80}
        & \graycell{6.40}
        & \graycell{8.70}
        & \graycell{3.20}
        & \graycell{11.00}
        & \graycell{42.10}
        & \graycell{12.50}
        & \graycell{27.20}
        & \graycell{14.10}
        & \graycell{14.50}
        \\
        GaussTR~\cite{jiang2025gausstr}
        & 1
        & 13.25
        & 2.09
        & 5.22
        & 14.07
        & 20.43
        & 5.70
        & 7.08
        & 5.12
        & 3.93
        & 0.92
        & 13.36
        & 39.44
        & 15.68
        & 22.89
        & 21.17
        & 21.87
        \\
        \graycell{PGOcc~\cite{yan2025progressive}}
        & \graycell{8}
        & \graycell{15.15}
        & \graycell{4.33}
        & \graycell{7.31}
        & \graycell{23.63}
        & \graycell{26.42}
        & \graycell{11.38}
        & \graycell{6.33}
        & \graycell{2.74}
        & \graycell{5.79}
        & \graycell{3.07}
        & \graycell{17.29}
        & \graycell{37.81}
        & \graycell{19.29}
        & \graycell{20.85}
        & \graycell{19.02}
        & \graycell{21.92}
        \\
        \midrule

        \modelname{}
        & 1
        & 14.19
        & 3.24
        & 4.83
        & 9.83
        & 27.93
        & 0.17
        & 5.26
        & 6.49
        & 5.48
        & 1.54
        & 13.64
        & 52.25
        & 15.78
        & 23.25
        & 21.74
        & 21.42
        \\

        \newmodelname{}
        & 1
        & 15.14
        & 3.73
        & 5.58
        & 10.11
        & 27.61
        & 2.52
        & 6.93
        & 7.37
        & 5.23
        & 1.51
        & 14.47
        & 55.42
        & 18.59
        & 26.32
        & 20.84
        & 20.93
        \\

        \bottomrule[1.2pt]
    \end{tabular}
    }
\end{table*}

\subsection{Datasets and Metrics}

\noindent\textbf{Indoor dataset.}
We train and evaluate on Occ-ScanNet~\cite{ISO}, where each sample provides a $60{\times}60{\times}36$ voxel grid covering a $4.8{\rm m}{\times}4.8{\rm m}{\times}2.88{\rm m}$ volume. The dataset contains 12 classes in total, including 11 semantic categories (\emph{ceiling}, \emph{floor}, \emph{wall}, \emph{window}, \emph{chair}, \emph{bed}, \emph{sofa}, \emph{table}, \emph{television}, \emph{furniture}, and \emph{objects}) and the \emph{empty} class.
We follow the standard split, consisting of 45{,}755 training samples and 19{,}764 validation samples.

\noindent\textbf{Outdoor dataset.}
To further evaluate the generalization of \newmodelname{} to large-scale autonomous-driving scenes, we conduct experiments on Occ3D-nuScenes~\cite{tian2023occ3d,nuscenes}.
Following the experimental protocol of PG-Occ~\cite{yan2025progressive}, Occ3D-nuScenes contains 1{,}000 driving scenes with dense semantic occupancy annotations derived from the nuScenes benchmark.
We consider a region of interest spanning $[-40,40]\,{\rm m}$ along both the $x$- and $y$-axes and $[-1,5.4]\,{\rm m}$ along the $z$-axis. The region is discretized into a $200{\times}200{\times}16$ voxel grid with a voxel size of $0.4\,{\rm m}$.

\noindent\textbf{Task setup.}
For Occ-ScanNet, we consider the \emph{monocular} setting, where a single RGB image is provided as input.
Following LegoOcc, we use binary occupancy labels to supervise 3D geometry.
For Occ3D-nuScenes, we adopt a \emph{single-frame} setting at inference, using only the six surround-view images from the current timestamp without temporal observations.
Following PG-Occ~\cite{yan2025progressive}, we train \textbf{without occupancy annotations}: 3D geometry is learned by Gaussian rendering across neighboring frames, while semantic supervision is obtained from open-vocabulary segmentation.
Importantly, neighboring frames are used only to provide supervision during training and are not required at inference, where our model operates on a single frame~\cite{gan2025gaussianocc,jiang2025gausstr}.

\noindent\textbf{Metrics.}
For Occ-ScanNet, following prior work~\cite{ISO,embodiedocc}, evaluation is restricted to the current camera frustum.
We report the overall occupancy IoU, per-class IoU over the 11 semantic categories, and their mean IoU (mIoU).
For Occ3D-nuScenes, following PG-Occ~\cite{yan2025progressive}, we report per-class IoU and mIoU over the 15 evaluated semantic categories.

\subsection{Implementation Details}

We adopt Depth-Anything V2~\cite{depthanythingv2} as our depth backbone, which has been used as a geometry prior in~\cite{ISO,embodiedocc}. Building on this, we construct Gaussian primitives using a surface-point expansion strategy similar to~\cite{gpocc}.
For semantics, we use Trident~\cite{trident} as the open-vocabulary segmentation model to extract language-aligned per-pixel features, and the CLIP~\cite{clip} text encoder to embed category names. We follow the standard prompt template ``a photo of $\{\cdot\}$'' to obtain text embeddings.

\noindent\textbf{Indoor implementation.}
All models are trained for 10 epochs using AdamW~\cite{adamw} with a global batch size of 8. The learning rate is initialized to $2\times10^{-4}$ and decayed with a cosine schedule~\cite{coslr} after linear warmup, with weight decay set to $0.01$.
We apply gradient clipping~\cite{gradclip} with a maximum norm of $1.0$. Input images are resized such that the longer side is 518 pixels while preserving the aspect ratio. All experiments are conducted on NVIDIA GeForce RTX 4090 GPUs.

\noindent\textbf{Outdoor implementation.}
For Occ3D-nuScenes~\cite{occ3d}, we follow PG-Occ~\cite{yan2025progressive} for the outdoor open-vocabulary setting and adopt the same training hyperparameters. Specifically, the model is trained on NVIDIA RTX 4090 GPUs using AdamW with a total batch size of 8, an initial learning rate of $1\times10^{-4}$, a weight decay of $0.01$, and gradient clipping with a maximum $\ell_2$ norm of $35$. We resize the input images to $336\times602$.

\subsection{Main Results}
\label{sec:main_results}

To assess the effectiveness of our approach, we conduct comprehensive evaluations on the indoor Occ-ScanNet benchmark~\cite{ISO} and the outdoor Occ3D-nuScenes benchmark~\cite{occ3d}.
For Occ-ScanNet, we report overall occupancy IoU together with per-class IoU and mean IoU (mIoU) in \Cref{tab:main_mono}.
For Occ3D-nuScenes, all methods are evaluated under the open-vocabulary setting without \textbf{using occupancy annotations during training}.
For clarity, rows in \textcolor[HTML]{2171B5}{blue} denote \colorbox{Sbg}{\textcolor[HTML]{2171B5}{closed-vocabulary}} (full-annotation) settings, whereas rows in \textcolor[HTML]{C05A00}{orange} denote \colorbox{Ubg}{\textcolor[HTML]{C05A00}{open-vocabulary}} (geometry-only annotation); darker shades highlight our methods.
For Occ3D-nuScenes, we report mean IoU (mIoU) and per-class IoU in \Cref{tab:main_nusc}.

\noindent\textbf{Indoor results on Occ-ScanNet.}
From~\Cref{tab:main_mono}, existing geometry-only open-vocabulary baselines remain substantially behind fully supervised occupancy models.
POP-3D~\cite{pop3d} and LOcc~\cite{yu2025language} obtain only 35.32/5.96 and 36.70/9.25 IoU/mIoU, respectively, showing the difficulty of recovering accurate 3D semantics by directly lifting 2D open-vocabulary features without semantic voxel supervision.
In contrast, \modelname{} achieves 59.50 IoU and 21.05 mIoU, substantially improving both geometric occupancy and semantic prediction under the same geometry-only annotation setting.
Notably, its occupancy IoU already exceeds all closed-vocabulary methods.
\newmodelname{} further improves the performance to 61.73 IoU and 24.22 mIoU, corresponding to gains of +2.23 IoU and +3.17 mIoU over \modelname{}.
The improvement in both metrics shows that the journal-version extensions enhance semantic prediction while further strengthening geometric reconstruction.
The per-class results exhibit particularly clear gains on ceiling (+13.32), sofa (+4.79), chair (+4.55), bed (+3.12), television (+3.15), and window (+3.06).
These categories typically have well-defined object or surface extents, for which covisibility-aware support-view selection and SAM-based instance consistency can provide cleaner cross-view supervision and stronger object-level semantic coherence.
Overall, these results demonstrate that \newmodelname{} substantially strengthens open-vocabulary semantic occupancy prediction while preserving the strong geometric accuracy of \modelname{}.

\noindent\textbf{Outdoor results on Occ3D-nuScenes.}
\Cref{tab:main_nusc} reports the results on the outdoor Occ3D-nuScenes benchmark.
We compare against representative camera-based occupancy methods that do not rely on LiDAR supervision, including both single-frame and temporal approaches.
Among previous single-frame methods, GaussTR~\cite{jiang2025gausstr} achieves the strongest mIoU of 13.25, while temporal methods such as PGOcc~\cite{yan2025progressive} use 8 frames and obtain 15.15 mIoU.
Using only a single frame, \modelname{} achieves 14.19 mIoU, outperforming all previous single-frame baselines by a clear margin, including GaussTR by +0.94 mIoU.
\newmodelname{} further raises the performance to 15.14 mIoU, yielding an additional +0.95 mIoU improvement.
Remarkably, this result nearly matches the 15.15 mIoU of PGOcc, despite PGOcc requiring eight temporal frames at inference.
The improvement of \newmodelname{} is distributed across both foreground objects and background surfaces.
Compared with \modelname{}, it notably improves construction vehicles (+2.35), motorcycles (+1.67), drivable surfaces (+3.17), sidewalks (+2.81), and terrain (+3.07).
These gains suggest that the proposed extensions improve not only object-level semantic discrimination but also the spatial consistency of large-scale scene regions.
Although small regressions are observed for several categories, such as car, traffic cone, manmade, and vegetation, the overall mIoU consistently increases.

The outdoor evaluation demonstrates that our approach generalizes effectively from indoor monocular scenes to large-scale autonomous-driving environments, while \newmodelname{} consistently improves performance across substantially different scene layouts and semantic distributions.

\subsection{Qualitative Visualization}

\begin{figure*}[h]
  \centering
  \includegraphics[width=0.98\textwidth]{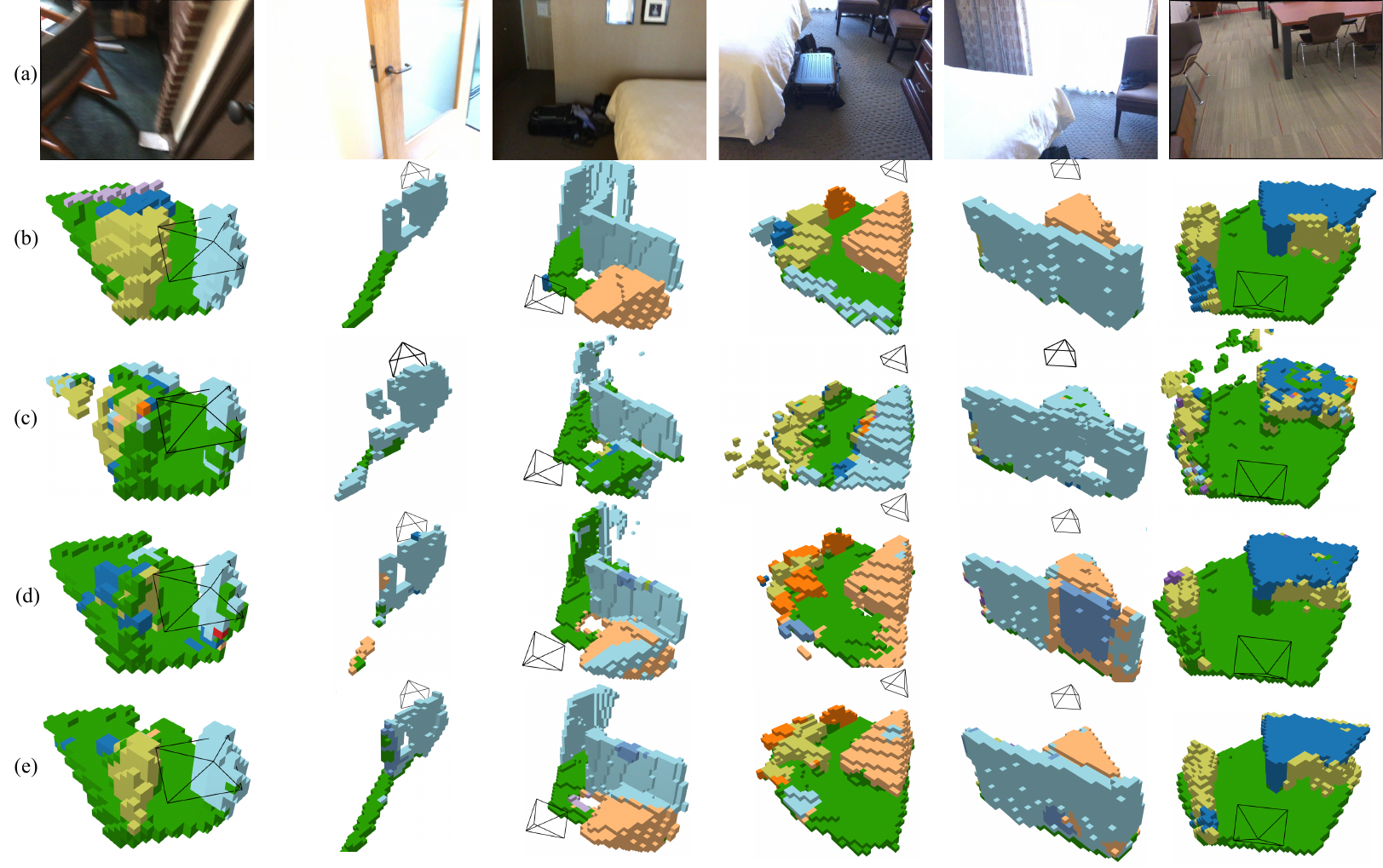}
  \vspace{-1mm}
  \caption{
\textbf{Qualitative results on Occ-ScanNet.}
From top to bottom:
(a) input images;
(b) ground-truth semantic occupancy;
(c) results from our re-implemented LOcc~\cite{yu2025language};
(d) \modelname{};
and (e) \newmodelname{}.
All compared open-vocabulary methods are trained with geometry-only supervision and evaluated using the closed-vocabulary annotations of Occ-ScanNet.
}
  \vspace{-1mm}
  \label{fig:quan_sl}
\end{figure*}

\begin{figure*}[h]
  \centering
  \includegraphics[width=0.96\textwidth]{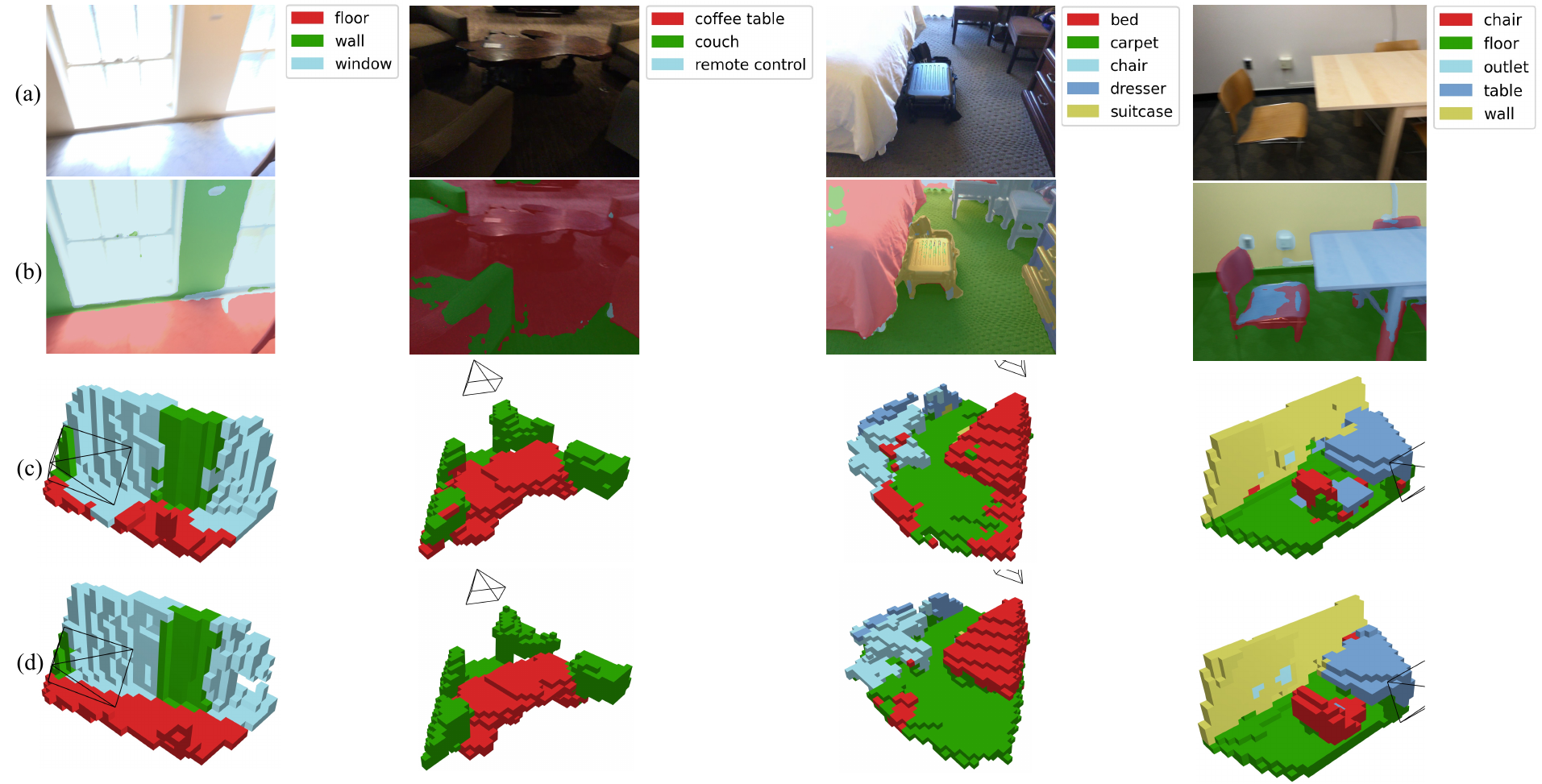}
  \vspace{-2mm}
  \caption{\textbf{Open-vocabulary qualitative comparison.} Legends list the \emph{VLM-extracted object nouns} used as text queries. (a) Input image. (b) Open-vocabulary 2D segmentation for the queried nouns. (c) \modelname~3D open-vocabulary occupancy, colored by the same categories. (d) \newmodelname~3D open-vocabulary occupancy, colored by the same categories.}
  \label{fig:quan_ov}
\end{figure*}

We present qualitative results under both closed and open vocabulary settings. As shown in~\Cref{fig:quan_sl}, our open-vocabulary model is evaluated against the closed-set semantics of Occ-ScanNet~\cite{ISO}. Across diverse scenes, the predicted occupancy aligns closely with ground-truth labels, indicating strong 
geometric accuracy and semantic generalization to the dataset taxonomy.
Beyond this closed set, \Cref{fig:quan_ov} presents open-vocabulary predictions obtained by
querying arbitrary given categories. For each image, a VLM~\cite{Qwen2.5-VL} extracts object \emph{nouns} as text queries, which our model uses to produce the corresponding 3D semantic occupancies.
More qualitative visualizations are provided in our appendix, including text-conditioned 3D grounding examples.
These results demonstrate the model's ability to localize and identify free-form indoor categories in 3D.

\noindent\textbf{Outdoor visualization.}
\Cref{fig:nusc_qual} presents qualitative comparisons on Occ3D-nuScenes.
Compared with GaussTR, \newmodelname{} better preserves the overall road layout and surrounding scene structure, while recovering substantially more complete long-range geometry.
The resulting semantic occupancy is visually more consistent with the ground truth, further demonstrating the applicability of our framework to large-scale outdoor driving scenes.

\begin{figure*}[h]
  \centering
  \includegraphics[width=0.98\textwidth]{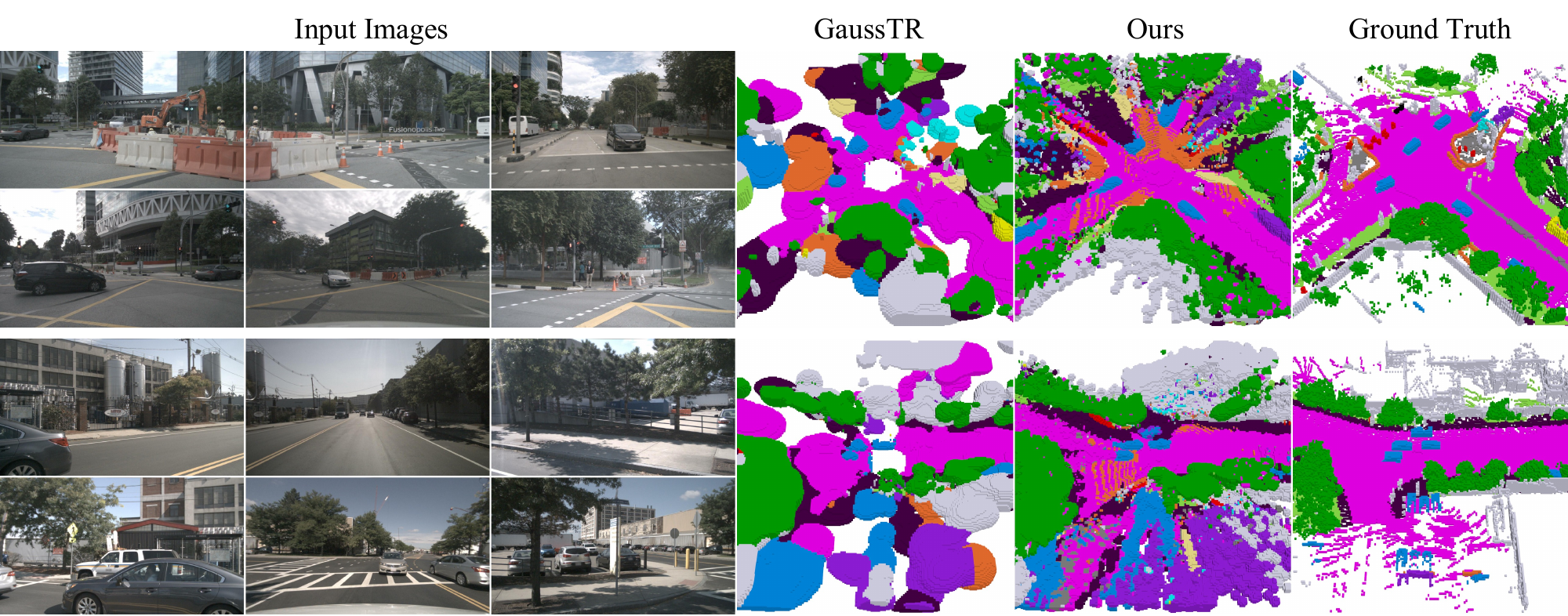}
  \vspace{-1mm}
  \caption{
  \textbf{Qualitative results on Occ3D-nuScenes.} We show surround-view input images and semantic occupancy predictions from GaussTR and \newmodelname{}, together with the ground truth. \newmodelname{} better preserves the overall road layout and long-range scene structure, yielding more complete occupancy predictions.}
  \label{fig:nusc_qual}
  \vspace{-1mm}
  \label{fig:nusc_qual}
\end{figure*}

\subsection{Ablation Studies}
We use the same coloring convention as in the main results for \colorbox{Sbg}{\textcolor[HTML]{2171B5}{closed-vocabulary}} and \colorbox{Ubg}{\textcolor[HTML]{C05A00}{open-vocabulary}} settings. When a table reports \emph{only} open-vocabulary results, background coloring is omitted for brevity. More ablations and analyses are provided in our appendix.

\begin{table}[h]\small
\centering
\vspace{-1mm}
\caption{\textbf{Ablation on the Gaussian-to-Occupancy operator.}
We compare three aggregation rules: GaussianFormer2, Bernoulli, and Poisson, under both \emph{closed-} and \emph{open-} vocabulary settings.}
\begin{tabular}{c|cc}
\toprule
Operation & mIoU & IoU \\
\midrule
\rowcolor{Sbg} GaussianFormer2 & 51.88 & 56.96 \\
\rowcolor{Sbg} Bernoulli & 51.68 & 58.06 \\
\rowcolor{SbgStrong} Poisson & 52.80 & 59.24 \\
\midrule
\rowcolor{Ubg} GaussianFormer2 & 0.00 & 0.00 \\
\rowcolor{Ubg} Bernoulli & 17.25 & 46.65 \\
\rowcolor{UbgStrong} Poisson & 21.05 & 59.50 \\
\bottomrule
\end{tabular}
\label{tab:abs_g2o}
\vspace{-1mm}
\end{table}

\noindent\textbf{Ablation on Gaussian-to-Occupancy (G2O).}
As shown in~\Cref{tab:abs_g2o}, replacing the operator from GaussianFormer2~\cite{gaussianformer2} with our Poisson-based G2O consistently improves performance under the \emph{closed-vocabulary} setting, 56.96$\!\rightarrow$59.24 IoU and 51.88$\!\rightarrow$52.80 mIoU.
The advantage is more pronounced in the \emph{open-vocabulary} case, where our Poisson G2O achieves 59.50 IoU and 21.05 mIoU, surpassing the Bernoulli variant by +12.85 IoU and +3.80 mIoU.
These results indicate that incorporating opacity via a Poisson-based compositing framework yields faithful voxel aggregation.

\begin{table}[h]\scriptsize
\centering
\caption{\textbf{Ablation on temperature scheduling.}
We vary the temperature range $(T_{\min}, T_{\max})$, the test-time temperature $\tau_\text{test}$, and the scheduling strategy.
“$\times$” denotes no schedule. %
}
\vspace{-1mm}
\label{tab:abs_temp}
\begin{tabular}{cccc|cc}
\toprule
$T_{min}$ & $T_{max}$ & $\tau_\text{test}$ & scheduler & mIoU & IoU \\
\midrule
1.0 & 1.0 & 1 & $\times$ & 18.15 & 59.19 \\
1.0 & 1.0 & 1e-3 & $\times$ & 12.83 & 32.52 \\
1e-3 & 1e-3 & 1e-3 & $\times$ & 0.00 & 0.00 \\
\midrule
1e-3 & 1.0 & 1e-3 & linear & 2.30 & 7.60 \\
\midrule
1e-2 & 1.0 & 1e-2 & exp & 20.85 & 59.25 \\
1e-3 & 1.0 & 1e-3 & exp & 21.05 & 59.50 \\
1e-4 & 1.0 & 1e-4 & exp & 20.37 & 58.54 \\
\bottomrule
\vspace{-5mm}
\end{tabular}
\end{table}

\noindent\textbf{Ablation on temperature scheduling.}
As reported in \Cref{tab:abs_temp}, fixing $\tau{=}1$ throughout training and testing (\ie, no schedule) yields reasonable geometry (59.19 IoU) but poor semantics (18.15 mIoU), indicating substantial per-ray feature mixing degrades semantic discrimination.
Using mismatched temperatures ($\tau_{\text{train}}{=}1$, $\tau_{\text{test}}{=}10^{-3}$)
severely degrades geometry (32.52 IoU) due to a train–test distribution shift. Keeping a constant low temperature ($\tau{=}10^{-3}$) collapses optimization entirely, underscoring the necessity of progressive decay.
A linear decay from $1\!\to\!10^{-3}$ with few iterations near $\tau\!\approx\!0$ also underperforms (2.30 mIoU/7.60 IoU), suggesting that allocating too few steps in the low-temperature regime inadequately sharpen per-Gaussian opacities.
In contrast, our exponential schedule, which allocates more iterations to the low-temperature regime, achieves the best results: 21.05 mIoU/59.50 IoU with $(T_{\min},T_{\max}){=}(10^{-3},1)$ and $\tau_{\text{test}}{=}10^{-3}$.
Lowering $T_{\min}$ to $10^{-4}$ slightly hurts performance (20.37 mIoU/58.54 IoU), suggesting over-sharpening amplifies noise. 
These results demonstrate that progressive temperature decay effectively reduces feature mixing and yields sharper, more discriminative Gaussian embeddings.

\begin{table}[h]\small
\centering
\caption{\textbf{Comparison of \modelname{} and \newmodelname{}.}
Starting from \modelname{} baseline, we progressively introduce covisibility-aware support-view selection (Covis.) and SAM-based instance-level semantic consistency (SAM-Inst.).}
\label{tab:abs_lego_plus_components}
\begin{tabular}{lcc|cc}
\toprule
Method & Covis. & SAM-Inst. & mIoU & IoU \\
\midrule
\modelname{} &  &  & 21.05 & 59.50 \\
\modelname{} + Covis. & $\checkmark$ & & 23.17 & 60.31 \\
\newmodelname{} & $\checkmark$ & $\checkmark$ & 24.22 & 61.73 \\
\bottomrule
\end{tabular}
\end{table}

\noindent\textbf{Comparison between \modelname{} and \newmodelname{}.}
\Cref{tab:abs_lego_plus_components} evaluates the two newly introduced components on Occ-ScanNet.
Starting from the \modelname{} baseline, covisibility-aware support-view selection improves mIoU/IoU from 21.05/59.50 to 23.17/60.31. This shows that selecting appropriate target views is important for rendering-based supervision, as sufficient geometric overlap yields more reliable feature alignment while avoiding weakly overlapping or redundant views.
Adding SAM-based instance-level semantic consistency further improves the result to 24.22/61.73, indicating that instance-level prototypes effectively complement pixel-wise alignment by reducing semantic inconsistency within the same object.
Overall, \newmodelname{} achieves gains of +3.17 mIoU and +2.23 IoU over the \modelname{} baseline.

\begin{table}[h!]\small
\caption{\textbf{Model profile.} FPS is measured on the same machine with a single RTX 4090 GPU, averaged over 1{,}000 runs after 100 warm-up runs.}
\setlength{\tabcolsep}{0.008\textwidth}
\captionsetup{font=scriptsize}
\centering
\begin{tabular}{lcccc}
\toprule
\textbf{Model} & mIoU & IoU & FPS & Params \\
\midrule
\rowcolor{Sbg} ISO~\cite{ISO} & 28.71 & 42.16 & 3.81 & 303.05M \\
\rowcolor{Sbg} EmbodiedOcc~\cite{embodiedocc} & 45.15 & 53.55 & 11.48 & 231.45M \\
\midrule
\rowcolor{Ubg} POP-3D~\cite{pop3d} & 5.96 & 35.32 & 10.21 & 115.31M \\
\rowcolor{Ubg} LOcc~\cite{yu2025language} & 9.25 & 36.70 & 8.93 & 113.50M \\
\rowcolor{UbgStrong} \modelname{} & 21.05 & 59.50 & 22.47 & 99.04M \\
\rowcolor{UbgStrong} \newmodelname{} & 24.22 & 61.73 & 22.47 & 99.04M \\
\bottomrule
\end{tabular}
\label{tab:fps}
\end{table}

\noindent\textbf{Model profile on Occ-ScanNet.}
We summarize the model profile in~\Cref{tab:fps}. ISO~\cite{ISO} and EmbodiedOcc~\cite{embodiedocc} rely on closed-set classification heads, whereas open-vocabulary methods perform semantic prediction by comparing occupancy features with language embeddings~\cite{clip} using cosine similarity.
\newmodelname{} retains the same parameter count and inference speed as \modelname{}. Consequently, the improvements of +3.17 mIoU and +2.23 IoU are achieved without additional inference overhead.

\begin{table}[h!]
\small
\caption{\textbf{Model profile.}
FPS is measured on the same machine using a single RTX 4090 GPU.}
\vspace{-1mm}
\setlength{\tabcolsep}{0.008\textwidth}
\centering
\begin{tabular}{lcccc}
\toprule
\textbf{Model} & \textbf{Frames} & \textbf{mIoU} &
\textbf{FPS} & \textbf{Params (Core / E2E)} \\
\midrule
GaussTR~\cite{jiang2025gausstr}
    & 1 & 13.25 & 0.42 & 13.74M / 575.59M \\
PG-Occ~\cite{yan2025progressive}
    & 8 & 15.15 & 0.24 & 90.03M / 501.97M \\
\midrule
\newmodelname{}
    & 1 & 15.14 & 6.22 & 161.58M / 161.58M \\
\bottomrule
\end{tabular}
\label{tab:nusc_fps}
\end{table}

\noindent\textbf{Model profile on nuScenes.}
We profile all methods on the same machine with a single NVIDIA RTX 4090 GPU and batch size 1. For parameter counting, we report both the occupancy core and the complete end-to-end inference pipeline, including auxiliary networks required at test time. GaussTR~\cite{jiang2025gausstr} requires FeatUp~\cite{featup} and Metric3D~\cite{yin2023metric3d}, resulting in 575.59M end-to-end parameters, while PG-Occ requires Metric3D and reaches 501.97M parameters. In contrast, \newmodelname{} integrates DepthAnything-V2~\cite{depthanythingv2} into the model and requires no additional network at inference, resulting in 161.58M parameters for both the core and the complete pipeline. As shown in \Cref{tab:nusc_fps}, \newmodelname{} achieves 6.22 FPS, approximately $15\times$ faster than GaussTR (0.42 FPS) and $26\times$ faster than the eight-frame PG-Occ pipeline (0.24 FPS), while maintaining competitive occupancy accuracy. Moreover, its complete inference pipeline uses 71.9\% fewer parameters than GaussTR and 67.8\% fewer than PG-Occ, demonstrating that \newmodelname{} offers a substantially more efficient end-to-end inference pipeline without relying on separately executed foundation models.

\section{Limitations}

Although \newmodelname{} demonstrates strong performance across indoor and outdoor benchmarks, several limitations remain. First, broad categories such as \emph{furniture} and \emph{objects} remain challenging because they contain visually and geometrically diverse instances, making language-aligned representations difficult to calibrate consistently. Second, while our outdoor formulation follows prior work and can be trained without occupancy annotations, the indoor setting still relies on binary occupancy supervision for reliable 3D geometry. Fully removing explicit geometry supervision in cluttered indoor environments through stronger multi-view geometric priors or self-supervised reconstruction objectives remains an important direction for future work.

\section{Conclusion}
We proposed \newmodelname{}, a general open-vocabulary 3D occupancy framework for indoor embodied environments and outdoor autonomous-driving scenarios. Building upon \modelname{}, our framework retains the opacity-aware Poisson-based Gaussian-to-Occupancy formulation and Progressive Temperature Decay for reliable geometry learning and language alignment, while introducing Instance-Level Semantic Consistency to reduce intra-object semantic fragmentation and Covisibility-Aware Support-View Selection for more informative multi-view supervision. Extensive experiments on Occ-ScanNet and Occ3D-nuScenes demonstrate strong performance across diverse scene domains and supervision settings, including geometry-only supervision for indoor scenes and annotation-free training for outdoor scenarios. We hope this work advances general and scalable open-vocabulary 3D scene understanding for embodied and autonomous systems.

\section*{Acknowledgment}

This work was supported in part by the National Natural Science Foundation
of China under Grant 62573370 and in part by the Education Department of
Guangdong Province under Grant 2025ZDZX3051.

\FloatBarrier

{
    \bibliographystyle{IEEEtran}
    \bibliography{main}

@String(CVPR= {IEEE Conf. Comput. Vis. Pattern Recog.})

@String(ICCV= {Int. Conf. Comput. Vis.})

@String(AAAI = {AAAI})

@String(CVPR  = {CVPR})

@String(ICCV  = {ICCV})

@article{pop3d,
  title={Pop-3d: Open-vocabulary 3d occupancy prediction from images},
  author={Vobecky, Antonin and Sim{\'e}oni, Oriane and Hurych, David and Gidaris, Spyridon and Bursuc, Andrei and P{\'e}rez, Patrick and Sivic, Josef},
  journal={Advances in Neural Information Processing Systems},
  volume={36},
  pages={50545--50557},
  year={2023}
}

@inproceedings{embodiedocc,
  title={Embodiedocc: Embodied 3d occupancy prediction for vision-based online scene understanding},
  author={Wu, Yuqi and Zheng, Wenzhao and Zuo, Sicheng and Huang, Yuanhui and Zhou, Jie and Lu, Jiwen},
  booktitle={2025 IEEE/CVF International Conference on Computer Vision (ICCV)},
  pages={26360--26370},
  year={2025},
  organization={IEEE}
}

@inproceedings{embodiedocc++,
  title={Embodiedocc++: Boosting embodied 3d occupancy prediction with plane regularization and uncertainty sampler},
  author={Wang, Hao and Wei, Xiaobao and Zhang, Xiaoan and Li, Jianing and Bai, Chengyu and Li, Ying and Lu, Ming and Zheng, Wenzhao and Zhang, Shanghang},
  booktitle={Proceedings of the 33rd ACM International Conference on Multimedia},
  pages={925--934},
  year={2025}
}

@inproceedings{gaussianformer,
  title={Gaussianformer: Scene as gaussians for vision-based 3d semantic occupancy prediction},
  author={Huang, Yuanhui and Zheng, Wenzhao and Zhang, Yunpeng and Zhou, Jie and Lu, Jiwen},
  booktitle={European Conference on Computer Vision},
  pages={376--393},
  year={2024},
  organization={Springer}
}

@inproceedings{gaussianformer2,
  title={Gaussianformer-2: Probabilistic gaussian superposition for efficient 3d occupancy prediction},
  author={Huang, Yuanhui and Thammatadatrakoon, Amonnut and Zheng, Wenzhao and Zhang, Yunpeng and Du, Dalong and Lu, Jiwen},
  booktitle={2025 IEEE/CVF Conference on Computer Vision and Pattern Recognition (CVPR)},
  pages={27477--27486},
  year={2025},
  organization={IEEE}
}

@inproceedings{monoscene,
  title={Monoscene: Monocular 3d semantic scene completion},
  author={Cao, Anh-Quan and De Charette, Raoul},
  booktitle={Proceedings of the IEEE/CVF Conference on Computer Vision and Pattern Recognition},
  pages={3991--4001},
  year={2022}
}

@inproceedings{ISO,
  title={Monocular occupancy prediction for scalable indoor scenes},
  author={Yu, Hongxiao and Wang, Yuqi and Chen, Yuntao and Zhang, Zhaoxiang},
  booktitle={European Conference on Computer Vision},
  pages={38--54},
  year={2024},
  organization={Springer}
}

@inproceedings{surroundocc,
  title={Surroundocc: Multi-camera 3d occupancy prediction for autonomous driving},
  author={Wei, Yi and Zhao, Linqing and Zheng, Wenzhao and Zhu, Zheng and Zhou, Jie and Lu, Jiwen},
  booktitle={Proceedings of the IEEE/CVF International Conference on Computer Vision},
  pages={21729--21740},
  year={2023}
}

@inproceedings{volumetrivln,
  title={Volumetric environment representation for vision-language navigation},
  author={Liu, Rui and Wang, Wenguan and Yang, Yi},
  booktitle={Proceedings of the IEEE/CVF conference on computer vision and pattern recognition},
  pages={16317--16328},
  year={2024}
}

@article{depthanythingv2,
  title={Depth anything v2},
  author={Yang, Lihe and Kang, Bingyi and Huang, Zilong and Zhao, Zhen and Xu, Xiaogang and Feng, Jiashi and Zhao, Hengshuang},
  journal={Advances in Neural Information Processing Systems},
  volume={37},
  pages={21875--21911},
  year={2024}
}

@inproceedings{voxformer,
  title={Voxformer: Sparse voxel transformer for camera-based 3d semantic scene completion},
  author={Li, Yiming and Yu, Zhiding and Choy, Christopher and Xiao, Chaowei and Alvarez, Jose M and Fidler, Sanja and Feng, Chen and Anandkumar, Anima},
  booktitle={Proceedings of the IEEE/CVF conference on computer vision and pattern recognition},
  pages={9087--9098},
  year={2023}
}

@article{occvla,
  title={OccVLA: Vision-Language-Action Model with Implicit 3D Occupancy Supervision},
  author={Liu, Ruixun and Kong, Lingyu and Li, Derun and Zhao, Hang},
  journal={arXiv preprint arXiv:2509.05578},
  year={2025}
}

@article{occllama,
  title={Occllama: An occupancy-language-action generative world model for autonomous driving},
  author={Wei, Julong and Yuan, Shanshuai and Li, Pengfei and Hu, Qingda and Gan, Zhongxue and Ding, Wenchao},
  journal={arXiv preprint arXiv:2409.03272},
  year={2024}
}

@inproceedings{occupancypoints,
  title={Occupancy as set of points},
  author={Shi, Yiang and Cheng, Tianheng and Zhang, Qian and Liu, Wenyu and Wang, Xinggang},
  booktitle={European Conference on Computer Vision},
  pages={72--87},
  year={2024},
  organization={Springer}
}

@inproceedings{lss,
  title={Lift, splat, shoot: Encoding images from arbitrary camera rigs by implicitly unprojecting to 3d},
  author={Philion, Jonah and Fidler, Sanja},
  booktitle={European conference on computer vision},
  pages={194--210},
  year={2020},
  organization={Springer}
}

@article{adamw,
  title={Decoupled weight decay regularization},
  author={Loshchilov, Ilya and Hutter, Frank},
  journal={arXiv preprint arXiv:1711.05101},
  year={2017}
}

@inproceedings{Triformer,
  title={Tri-perspective view for vision-based 3d semantic occupancy prediction},
  author={Huang, Yuanhui and Zheng, Wenzhao and Zhang, Yunpeng and Zhou, Jie and Lu, Jiwen},
  booktitle={Proceedings of the IEEE/CVF conference on computer vision and pattern recognition},
  pages={9223--9232},
  year={2023}
}

@inproceedings{nuscenes,
  title={nuscenes: A multimodal dataset for autonomous driving},
  author={Caesar, Holger and Bankiti, Varun and Lang, Alex H and Vora, Sourabh and Liong, Venice Erin and Xu, Qiang and Krishnan, Anush and Pan, Yu and Baldan, Giancarlo and Beijbom, Oscar},
  booktitle={Proceedings of the IEEE/CVF conference on computer vision and pattern recognition},
  pages={11621--11631},
  year={2020}
}

@inproceedings{occ3d,
  title={Occ3D: A Large-Scale 3D Occupancy Prediction Benchmark for Autonomous Driving},
  author={Tian, Xiaoyu and Jiang, Tao and Yun, Longfei and Mao, Yucheng and Yang, Huitong and Wang, Yue and Wang, Yilun and Zhao, Hang},
  booktitle={Advances in Neural Information Processing Systems},
  year={2023}
}

@inproceedings{ndcscene,
  title={Ndc-scene: Boost monocular 3d semantic scene completion in normalized device coordinates space},
  author={Yao, Jiawei and Li, Chuming and Sun, Keqiang and Cai, Yingjie and Li, Hao and Ouyang, Wanli and Li, Hongsheng},
  booktitle={2023 IEEE/CVF International Conference on Computer Vision (ICCV)},
  pages={9421--9431},
  year={2023},
  organization={IEEE Computer Society}
}

@inproceedings{sscnet,
  title={Semantic scene completion from a single depth image},
  author={Song, Shuran and Yu, Fisher and Zeng, Andy and Chang, Angel X and Savva, Manolis and Funkhouser, Thomas},
  booktitle={Proceedings of the IEEE conference on computer vision and pattern recognition},
  pages={1746--1754},
  year={2017}
}

@inproceedings{sparseocc,
  title={Sparseocc: Rethinking sparse latent representation for vision-based semantic occupancy prediction},
  author={Tang, Pin and Wang, Zhongdao and Wang, Guoqing and Zheng, Jilai and Ren, Xiangxuan and Feng, Bailan and Ma, Chao},
  booktitle={Proceedings of the IEEE/CVF Conference on Computer Vision and Pattern Recognition},
  pages={15035--15044},
  year={2024}
}

@article{fbocc,
  title={{FB-OCC}: {3D} Occupancy Prediction based on Forward-Backward View Transformation},
  author={Li, Zhiqi and Yu, Zhiding and Austin, David and Fang, Mingsheng and Lan, Shiyi and Kautz, Jan and Alvarez, Jose M},
  journal={arXiv:2307.01492},
  year={2023}
}

@article{gaussiansplatting,
  title={3D Gaussian splatting for real-time radiance field rendering.},
  author={Kerbl, Bernhard and Kopanas, Georgios and Leimk{\"u}hler, Thomas and Drettakis, George},
  journal={ACM Trans. Graph.},
  volume={42},
  number={4},
  pages={139--1},
  year={2023}
}

@article{gao2025loc,
  title={LOC: A General Language-Guided Framework for Open-Set 3D Occupancy Prediction},
  author={Gao, Yuhang and Xiang, Xiang and Zhong, Sheng and Wang, Guoyou},
  journal={arXiv preprint arXiv:2510.22141},
  year={2025}
}

@article{dong2023og,
  title={OG: Equip vision occupancy with instance segmentation and visual grounding},
  author={Dong, Zichao and Ji, Hang and Zhang, Weikun and Huang, Xufeng and Chen, Junbo},
  journal={arXiv preprint arXiv:2307.05873},
  year={2023}
}

@article{groundingocc,
  title={A Coarse-to-Fine Approach to Multi-Modality 3D Occupancy Grounding},
  author={Shi, Zhan and Wang, Song and Chen, Junbo and Zhu, Jianke},
  journal={arXiv preprint arXiv:2508.01197},
  year={2025}
}

@inproceedings{li2025ago,
  title={Ago: Adaptive grounding for open world 3d occupancy prediction},
  author={Li, Peizheng and Ding, Shuxiao and Zhou, You and Zhang, Qingwen and Inak, Onat and Triess, Larissa and Hanselmann, Niklas and Cordts, Marius and Zell, Andreas},
  booktitle={2025 IEEE/CVF International Conference on Computer Vision (ICCV)},
  pages={8645--8655},
  year={2025},
  organization={IEEE}
}

@article{occnerf,
  title={Occnerf: Advancing 3d occupancy prediction in lidar-free environments},
  author={Zhang, Chubin and Yan, Juncheng and Wei, Yi and Li, Jiaxin and Liu, Li and Tang, Yansong and Duan, Yueqi and Lu, Jiwen},
  journal={IEEE Transactions on Image Processing},
  volume={34},
  pages={3096--3107},
  year={2025},
  publisher={IEEE}
}

@article{boeder2024langocc,
  title={Langocc: Self-supervised open vocabulary occupancy estimation via volume rendering},
  author={Boeder, Simon and Gigengack, Fabian and Risse, Benjamin},
  journal={arXiv preprint arXiv:2407.17310},
  year={2024}
}

@inproceedings{yu2025language,
  title={Language driven occupancy prediction},
  author={Yu, Zhu and Pang, Bowen and Liu, Lizhe and Zhang, Runmin and Li, Qiang and Cao, Si-Yuan and Luo, Maochun and Chen, Mingxia and Yang, Sheng and Shen, Hui-Liang},
  booktitle={Proceedings of the IEEE/CVF International Conference on Computer Vision},
  pages={7548--7558},
  year={2025}
}

@inproceedings{veon,
  title={Veon: Vocabulary-enhanced occupancy prediction},
  author={Zheng, Jilai and Tang, Pin and Wang, Zhongdao and Wang, Guoqing and Ren, Xiangxuan and Feng, Bailan and Ma, Chao},
  booktitle={European Conference on Computer Vision},
  pages={92--108},
  year={2024},
  organization={Springer}
}

@inproceedings{clip,
  title={Learning transferable visual models from natural language supervision},
  author={Radford, Alec and Kim, Jong Wook and Hallacy, Chris and Ramesh, Aditya and Goh, Gabriel and Agarwal, Sandhini and Sastry, Girish and Askell, Amanda and Mishkin, Pamela and Clark, Jack and others},
  booktitle={International conference on machine learning},
  pages={8748--8763},
  year={2021},
  organization={PmLR}
}

@InProceedings{selfocc,
    author    = {Huang, Yuanhui and Zheng, Wenzhao and Zhang, Borui and Zhou, Jie and Lu, Jiwen},
    title     = {SelfOcc: Self-Supervised Vision-Based 3D Occupancy Prediction},
    booktitle = {Proceedings of the IEEE/CVF Conference on Computer Vision and Pattern Recognition (CVPR)},
    month     = {June},
    year      = {2024},
    pages     = {19946-19956}
}

@article{yu2025shtocc,
  title={SHTOcc: Effective 3D Occupancy Prediction with Sparse Head and Tail Voxels},
  author={Yu, Qiucheng and Xie, Yuan and Tan, Xin},
  journal={arXiv preprint arXiv:2505.22461},
  year={2025}
}

@InProceedings{OccFormer,
    author    = {Zhang, Yunpeng and Zhu, Zheng and Du, Dalong},
    title     = {OccFormer: Dual-path Transformer for Vision-based 3D Semantic Occupancy Prediction},
    booktitle = {Proceedings of the IEEE/CVF International Conference on Computer Vision (ICCV)},
    month     = {October},
    year      = {2023},
    pages     = {9433-9443}
}

@inproceedings{gaussrender,
  title={Gaussrender: Learning 3d occupancy with gaussian rendering},
  author={Chambon, Loick and Zablocki, Eloi and Boulch, Alexandre and Chen, Mickael and Cord, Matthieu},
  booktitle={Proceedings of the IEEE/CVF International Conference on Computer Vision},
  pages={27010--27020},
  year={2025}
}

@INPROCEEDINGS{scfusion,
  author={Wu, Shun-Cheng and Tateno, Kesuke and Navab, Nassir and Tombari, Federico},
  booktitle={2020 International Conference on 3D Vision (3DV)}, 
  title={SCFusion: Real-time Incremental Scene Reconstruction with Semantic Completion}, 
  year={2020},
  pages={801-810},
  doi={10.1109/3DV50981.2020.00090}}

@ARTICLE{supereight,
  author={Vespa, Emanuele and Nikolov, Nikolay and Grimm, Marius and Nardi, Luigi and Kelly, Paul H. J. and Leutenegger, Stefan},
  journal={IEEE Robotics and Automation Letters}, 
  title={Efficient Octree-Based Volumetric SLAM Supporting Signed-Distance and Occupancy Mapping}, 
  year={2018},
  volume={3},
  number={2},
  pages={1144-1151},
  doi={10.1109/LRA.2018.2792537}}

@inproceedings{forknet,
  title={Forknet: Multi-branch volumetric semantic completion from a single depth image},
  author={Wang, Yida and Tan, David Joseph and Navab, Nassir and Tombari, Federico},
  booktitle={2019 IEEE/CVF International Conference on Computer Vision (ICCV)},
  pages={8607--8616},
  year={2019},
  organization={IEEE}
}

@misc{ovo,
      title={OVO: Open-Vocabulary Occupancy}, 
      author={Zhiyu Tan and Zichao Dong and Cheng Zhang and Weikun Zhang and Hang Ji and Hao Li},
      year={2023},
      eprint={2305.16133},
      archivePrefix={arXiv},
      primaryClass={cs.CV},
      url={https://arxiv.org/abs/2305.16133}, 
}

@inproceedings{trident,
  title={Harnessing vision foundation models for high-performance, training-free open vocabulary segmentation},
  author={Shi, Yuheng and Dong, Minjing and Xu, Chang},
  booktitle={2025 IEEE/CVF International Conference on Computer Vision (ICCV)},
  pages={23487--23497},
  year={2025},
  organization={IEEE}
}

@inproceedings{drsplat25,
    title={Dr. Splat: Directly Referring 3D Gaussian Splatting via Direct Language Embedding Registration},
    author={Jun-Seong, Kim and Kim GeonU and Yu-Ji, Kim and Yu-Chiang Frank Wang and Jaesung Choe and Oh, Tae-Hyun},
    booktitle=CVPR,
    year={2025}
}

@inproceedings{groundingdino,
  title={Grounding dino: Marrying dino with grounded pre-training for open-set object detection},
  author={Liu, Shilong and Zeng, Zhaoyang and Ren, Tianhe and Li, Feng and Zhang, Hao and Yang, Jie and Jiang, Qing and Li, Chunyuan and Yang, Jianwei and Su, Hang and others},
  booktitle={European conference on computer vision},
  pages={38--55},
  year={2024},
  organization={Springer}
}

@article{groundedsam,
  title={Grounded sam: Assembling open-world models for diverse visual tasks},
  author={Ren, Tianhe and Liu, Shilong and Zeng, Ailing and Lin, Jing and Li, Kunchang and Cao, He and Chen, Jiayu and Huang, Xinyu and Chen, Yukang and Yan, Feng and others},
  journal={arXiv preprint arXiv:2401.14159},
  year={2024}
}

@article{Qwen2.5-VL,
  title={Qwen2.5-VL Technical Report},
  author={Bai, Shuai and Chen, Keqin and Liu, Xuejing and Wang, Jialin and Ge, Wenbin and Song, Sibo and Dang, Kai and Wang, Peng and Wang, Shijie and Tang, Jun and Zhong, Humen and Zhu, Yuanzhi and Yang, Mingkun and Li, Zhaohai and Wan, Jianqiang and Wang, Pengfei and Ding, Wei and Fu, Zheren and Xu, Yiheng and Ye, Jiabo and Zhang, Xi and Xie, Tianbao and Cheng, Zesen and Zhang, Hang and Yang, Zhibo and Xu, Haiyang and Lin, Junyang},
  journal={arXiv preprint arXiv:2502.13923},
  year={2025}
}

@inproceedings{OpenOcc,
  title={OpenOcc: Open Vocabulary 3D Scene Reconstruction via Occupancy Representation},
  author={Jiang, Haochen and Xu, Yueming and Zeng, Yihan and Xu, Hang and Zhang, Wei and Feng, Jianfeng and Zhang, Li},
  booktitle={IEEE/RSJ International Conference on Intelligent Robots and Systems (IROS)},
  year={2024},
}

@article{Yu2024GOF,
  author    = {Yu, Zehao and Sattler, Torsten and Geiger, Andreas},
  title     = {Gaussian Opacity Fields: Efficient Adaptive Surface Reconstruction in Unbounded Scenes},
  journal   = {ACM Transactions on Graphics},
  year      = {2024},
}

@inproceedings{qian2025splatsscdecoupleddepthguidedgaussian,
  title={Splatssc: Decoupled depth-guided gaussian splatting for semantic scene completion},
  author={Qian, Rui and Cao, Haozhi and Deng, Tianchen and Yuan, Shenghai and Xie, Lihua},
  booktitle={Proceedings of the AAAI Conference on Artificial Intelligence},
  volume={40},
  number={10},
  pages={8520--8528},
  year={2026}
}

@article{roboocc,
  title={Roboocc: Enhancing the geometric and semantic scene understanding for robots},
  author={Zhang, Zhang and Zhang, Qiang and Cui, Wei and Shi, Shuai and Guo, Yijie and Han, Gang and Zhao, Wen and Ren, Hengle and Xu, Renjing and Tang, Jian},
  journal={arXiv preprint arXiv:2504.14604},
  year={2025}
}

@inproceedings{gradclip,
  title={On the difficulty of training recurrent neural networks},
  author={Pascanu, Razvan and Mikolov, Tomas and Bengio, Yoshua},
  booktitle={International conference on machine learning},
  pages={1310--1318},
  year={2013},
  organization={Pmlr}
}

@InProceedings{scan2cad,
author = {Avetisyan, Armen and Dahnert, Manuel and Dai, Angela and Savva, Manolis and Chang, Angel X. and Niessner, Matthias},
title = {Scan2CAD: Learning CAD Model Alignment in RGB-D Scans},
booktitle = {The IEEE Conference on Computer Vision and Pattern Recognition (CVPR)},
month = {June},
year = {2019}
}

@article{zhang2023deeplongtailsurvey,
  title={Deep long-tailed learning: A survey},
  author={Zhang, Yifan and Kang, Bingyi and Hooi, Bryan and Yan, Shuicheng and Feng, Jiashi},
  journal={IEEE transactions on pattern analysis and machine intelligence},
  volume={45},
  number={9},
  pages={10795--10816},
  year={2023},
  publisher={IEEE}
}

@inproceedings{joseph2021towards,
  title={Towards open world object detection},
  author={Joseph, KJ and Khan, Salman and Khan, Fahad Shahbaz and Balasubramanian, Vineeth N},
  booktitle={Proceedings of the IEEE/CVF conference on computer vision and pattern recognition},
  pages={5830--5840},
  year={2021}
}

@inproceedings{shi2023legaussians,
  title={Language embedded 3d gaussians for open-vocabulary scene understanding},
  author={Shi, Jin-Chuan and Wang, Miao and Duan, Hao-Bin and Guan, Shao-Hua},
  booktitle={2024 IEEE/CVF Conference on Computer Vision and Pattern Recognition (CVPR)},
  pages={5333--5343},
  year={2024},
  organization={IEEE}
}

@inproceedings{zhou2024featuregs,
    title={Feature 3dgs: Supercharging 3d gaussian splatting to enable distilled feature fields},
    author={Zhou, Shijie and Chang, Haoran and Jiang, Sicheng and Fan, Zhiwen and Zhu, Zehao and Xu, Dejia and Chari, Pradyumna and You, Suya and Wang, Zhangyang and Kadambi, Achuta},
    booktitle={Proceedings of the IEEE/CVF Conference on Computer Vision and Pattern Recognition},
    pages={21676--21685},
    year={2024}
}

@book{kingman-poisson-processes,
  address = {New York},
  author = {Kingman, J. F. C.},
  description = {MR: Selected Matches for: Author=(Kingman) AND Title=(Poisson Processes)},
  isbn = {0-19-853693-3},
  mrclass = {60G05 (60G55 60K99)},
  mrnumber = {MR1207584 (94a:60052)},
  mrreviewer = {Dietrich Stoyan},
  note = {Oxford Science Publications},
  pages = {viii+104},
  publisher = {The Clarendon Press Oxford University Press},
  series = {Oxford Studies in Probability},
  title = {Poisson processes},
  volume = 3,
  year = 1993
}

@book{ross1996stochastic,
  author = {Ross, S.M.},
  isbn = {9780471120629},
  lccn = {95038012},
  publisher = {Wiley},
  series = {Wiley series in probability and statistics: Probability and statistics},
  title = {Stochastic processes},
  url = {https://books.google.de/books?id=ImUPAQAAMAAJ},
  year = 1996
}

@inproceedings{park2025s2gostreamingsparsegaussian,
 author = {Park, Jinhyung and Peng, Chensheng and hu, yihan and Zheng, Wenzhao and Kitani, Kris and Zhan, Wei},
 booktitle = {International Conference on Learning Representations},
 editor = {C. Vondrick and B. Hariharan and C. Raffel and L. Pinto and D. Yang and A. Faust},
 pages = {62911--62932},
 title = {S2GO: Streaming Sparse Gaussian Occupancy},
 url = {https://proceedings.iclr.cc/paper_files/paper/2026/file/669b0bb64ff9eff7a81a7858a54fe7a0-Paper-Conference.pdf},
 volume = {2026},
 year = {2026}
}

@inproceedings{gpocc,
  title={Generalizing Visual Geometry Priors to Sparse Gaussian Occupancy Prediction},
  author={Zhou, Changqing and Luo, Yueru and Chen, Changhao},
  booktitle={Proceedings of the IEEE/CVF Conference on Computer Vision and Pattern Recognition},
  pages={28578--28587},
  year={2026}
}

@article{gpoccpp,
  title={GPOcc++: Unified Sparse Gaussian Occupancy Prediction with Visual Geometry Priors},
  author={Zhou, Changqing and Luo, Yueru and Guo, Yulan and Wang, Bing and Qin, Jie and Chen, Changhao},
  journal={arXiv preprint arXiv:2607.13481},
  year={2026}
}

@inproceedings{legoocc,
  title={Monocular open vocabulary occupancy prediction for indoor scenes},
  author={Zhou, Changqing and Luo, Yueru and Zhang, Han and Jiang, Zeyu and Chen, Changhao},
  booktitle={Proceedings of the IEEE/CVF Conference on Computer Vision and Pattern Recognition},
  pages={21627--21637},
  year={2026}
}

@inproceedings{gan2025gaussianocc,
  title={Gaussianocc: Fully self-supervised and efficient 3d occupancy estimation with gaussian splatting},
  author={Gan, Wanshui and Liu, Fang and Xu, Hongbin and Mo, Ningkai and Yokoya, Naoto},
  booktitle={Proceedings of the IEEE/CVF International Conference on Computer Vision},
  pages={28980--28990},
  year={2025}
}

@inproceedings{jiang2025gausstr,
  title={Gausstr: Foundation model-aligned gaussian transformer for self-supervised 3d spatial understanding},
  author={Jiang, Haoyi and Liu, Liu and Cheng, Tianheng and Wang, Xinjie and Lin, Tianwei and Su, Zhizhong and Liu, Wenyu and Wang, Xinggang},
  booktitle={Proceedings of the IEEE/CVF Conference on Computer Vision and Pattern Recognition},
  pages={11960--11970},
  year={2025}
}

@inproceedings{yan2025progressive,
 author = {Yan, Chi and Xu, Dan},
 booktitle = {International Conference on Learning Representations},
 editor = {C. Vondrick and B. Hariharan and C. Raffel and L. Pinto and D. Yang and A. Faust},
 pages = {4715--4737},
 title = {Progressive Gaussian Transformer with Anisotropy-aware Sampling for Open Vocabulary Occupancy Prediction},
 url = {https://proceedings.iclr.cc/paper_files/paper/2026/file/08d2d3c98868539671e68b70679b8bff-Paper-Conference.pdf},
 volume = {2026},
 year = {2026}
}

@inproceedings{sam,
  title={Segment anything},
  author={Kirillov, Alexander and Mintun, Eric and Ravi, Nikhila and Mao, Hanzi and Rolland, Chloe and Gustafson, Laura and Xiao, Tete and Whitehead, Spencer and Berg, Alexander C and Lo, Wan-Yen and others},
  booktitle={Proceedings of the IEEE/CVF international conference on computer vision},
  pages={4015--4026},
  year={2023}
}

@inproceedings{featup,
  title={Featup: A model-agnostic framework for features at any resolution},
  author={Fu, Stephanie and Hamilton, Mark and Brandt, Laura E and Feldmann, Axel and Zhang, Zhoutong and Freeman, William},
  booktitle={International Conference on Learning Representations},
  volume={2024},
  pages={45324--45350},
  year={2024}
}

@inproceedings{yin2023metric3d,
  title={Metric3d: Towards zero-shot metric 3d prediction from a single image},
  author={Yin, Wei and Zhang, Chi and Chen, Hao and Cai, Zhipeng and Yu, Gang and Wang, Kaixuan and Chen, Xiaozhi and Shen, Chunhua},
  booktitle={2023 IEEE/CVF International Conference on Computer Vision (ICCV)},
  pages={9009--9019},
  year={2023},
  organization={IEEE}
}

@INPROCEEDINGS{jiang2026freeocc, 
    AUTHOR    = {Zeyu Jiang AND Changqing Zhou AND Xingxing Zuo AND Changhao Chen}, 
    TITLE     = {{FreeOcc: Training-Free Embodied Open-Vocabulary Occupancy Prediction}}, 
    BOOKTITLE = {Proceedings of Robotics: Science and Systems}, 
    YEAR      = {2026}, 
    ADDRESS   = {Sydney, Australia}, 
    MONTH     = {July}, 
    DOI       = {10.15607/RSS.2026.XXII.166} 
}

@article{li2026hierarchical,
  title={Hierarchical context alignment with disentangled geometric and temporal modeling for semantic occupancy prediction},
  author={Li, Bohan and Deng, Jiajun and Sun, Yasheng and Wang, Xiaofeng and Jin, Xin and Zeng, Wenjun},
  journal={IEEE Transactions on Pattern Analysis and Machine Intelligence},
  year={2026},
  publisher={IEEE}
}

@article{wei2026onebev,
  title={OneBEV++: Towards Unifying Bird's-Eye-View Semantic Mapping with Panoramas},
  author={Wei, Jiale and Teng, Zhifeng and Teng, Fei and Zheng, Junwei and Liu, Ruiping and Chen, Yufan and Hu, Jie and Yang, Kailun and Zhang, Jiaming and Stiefelhagen, Rainer},
  journal={IEEE Transactions on Pattern Analysis and Machine Intelligence},
  year={2026},
  publisher={IEEE}
}

@article{xie2025benchmarking,
  title={Benchmarking and improving bird’s eye view perception robustness in autonomous driving},
  author={Xie, Shaoyuan and Kong, Lingdong and Zhang, Wenwei and Ren, Jiawei and Pan, Liang and Chen, Kai and Liu, Ziwei},
  journal={IEEE Transactions on Pattern Analysis and Machine Intelligence},
  volume={47},
  number={5},
  pages={3878--3894},
  year={2025},
  publisher={IEEE}
}

@article{chen2025open,
  title={Open-crb: Toward open world active learning for 3d object detection},
  author={Chen, Zhuoxiao and Luo, Yadan and Wang, Zixin and Wang, Zijian and Huang, Zi},
  journal={IEEE Transactions on Pattern Analysis and Machine Intelligence},
  volume={47},
  number={10},
  pages={8336--8350},
  year={2025},
  publisher={IEEE}
}

@article{tan2024epmf,
  title={EPMF: Efficient perception-aware multi-sensor fusion for 3D semantic segmentation},
  author={Tan, Mingkui and Zhuang, Zhuangwei and Chen, Sitao and Li, Rong and Jia, Kui and Wang, Qicheng and Li, Yuanqing},
  journal={IEEE Transactions on Pattern Analysis and Machine Intelligence},
  volume={46},
  number={12},
  pages={8258--8273},
  year={2024},
  publisher={IEEE}
}

@article{hong2024unified,
  title={Unified 3d and 4d panoptic segmentation via dynamic shifting networks},
  author={Hong, Fangzhou and Kong, Lingdong and Zhou, Hui and Zhu, Xinge and Li, Hongsheng and Liu, Ziwei},
  journal={IEEE Transactions on Pattern Analysis and Machine Intelligence},
  volume={46},
  number={5},
  pages={3480--3495},
  year={2024},
  publisher={IEEE}
}

@article{piccinelli2025unidepthv2,
  title={Unidepthv2: Universal monocular metric depth estimation made simpler},
  author={Piccinelli, Luigi and Sakaridis, Christos and Yang, Yung-Hsu and Segu, Mattia and Li, Siyuan and Abbeloos, Wim and Van Gool, Luc},
  journal={IEEE Transactions on Pattern Analysis and Machine Intelligence},
  year={2025},
  publisher={IEEE}
}

@article{shao2024nddepth,
  title={Nddepth: Normal-distance assisted monocular depth estimation and completion},
  author={Shao, Shuwei and Pei, Zhongcai and Chen, Weihai and Chen, Peter CY and Li, Zhengguo},
  journal={IEEE Transactions on Pattern Analysis and Machine Intelligence},
  volume={46},
  number={12},
  pages={8883--8899},
  year={2024},
  publisher={IEEE}
}

@article{sarlin2023pixel,
  title={Pixel-perfect structure-from-motion with featuremetric refinement},
  author={Sarlin, Paul-Edouard and Lindenberger, Philipp and Larsson, Viktor and Pollefeys, Marc},
  journal={IEEE Transactions on Pattern Analysis and Machine Intelligence},
  volume={47},
  number={5},
  pages={3298--3309},
  year={2023},
  publisher={IEEE}
}

@article{liu2025explicit,
  title={Explicit visual prompting for universal foreground segmentations},
  author={Liu, Weihuang and Shen, Xi and Pun, Chi-Man and Cun, Xiaodong},
  journal={IEEE Transactions on Pattern Analysis and Machine Intelligence},
  year={2025},
  publisher={IEEE}
}

@article{chen2025constraint,
  title={Constraint-aware zero-shot vision-language navigation in continuous environments},
  author={Chen, Kehan and An, Dong and Huang, Yan and Xu, Rongtao and Su, Yifei and Ling, Yonggen and Reid, Ian and Wang, Liang},
  journal={IEEE Transactions on Pattern Analysis and Machine Intelligence},
  year={2025},
  publisher={IEEE}
}
}

\vspace{-8mm}

\begin{IEEEbiography}[{\includegraphics[width=1in,height=1.25in,clip,keepaspectratio]{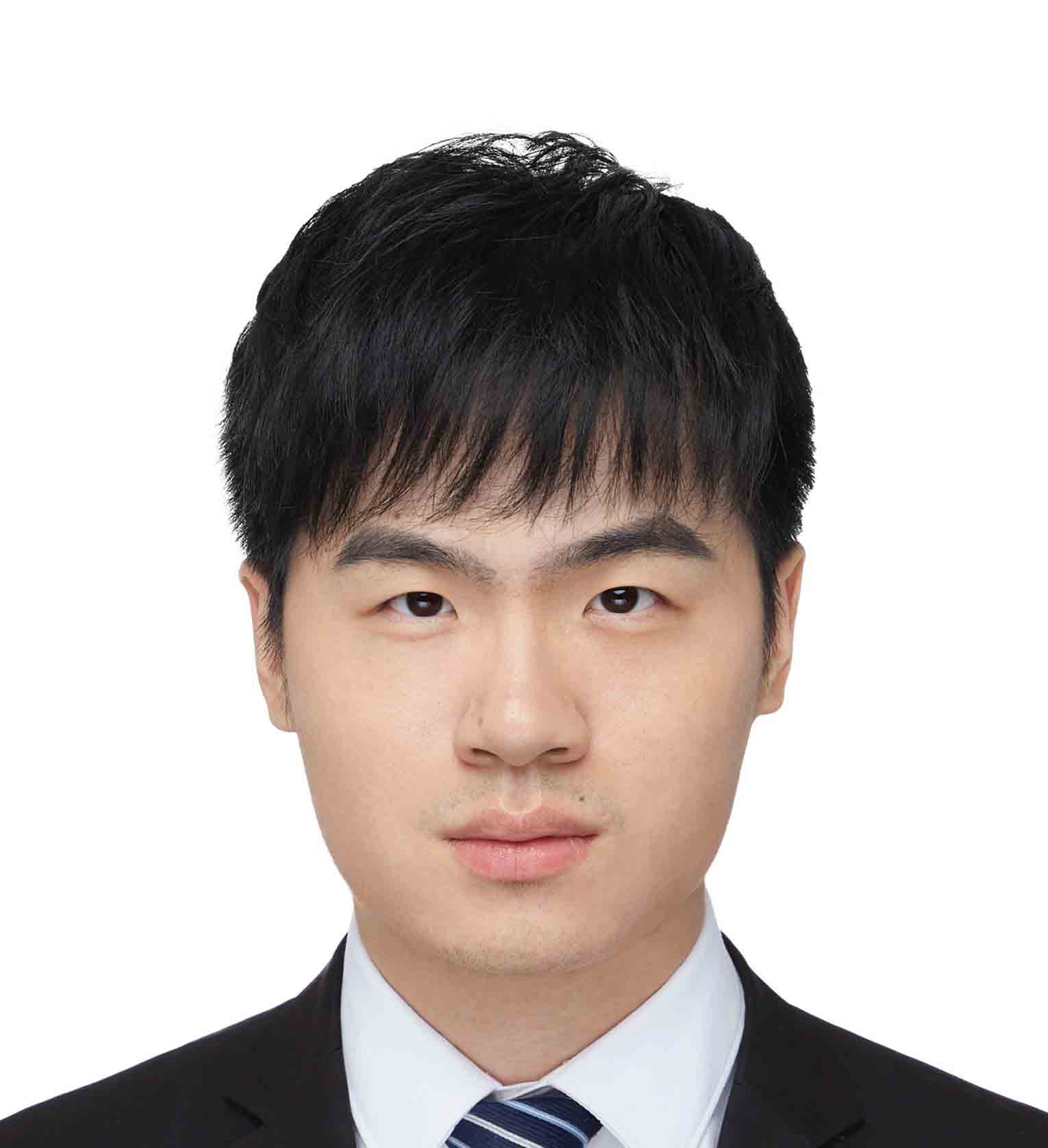}}]
{Changqing Zhou} is currently pursuing the Ph.D. degree at The Hong Kong University of Science and Technology (Guangzhou). His research interests include 3D computer vision, embodied scene understanding, occupancy prediction, and open-vocabulary 3D perception. His recent work focuses on 3D scene understanding and world modeling for visual navigation.
\end{IEEEbiography}

\vspace{-4mm}

\begin{IEEEbiography}[{\includegraphics[width=1in,height=1.25in,clip,keepaspectratio]{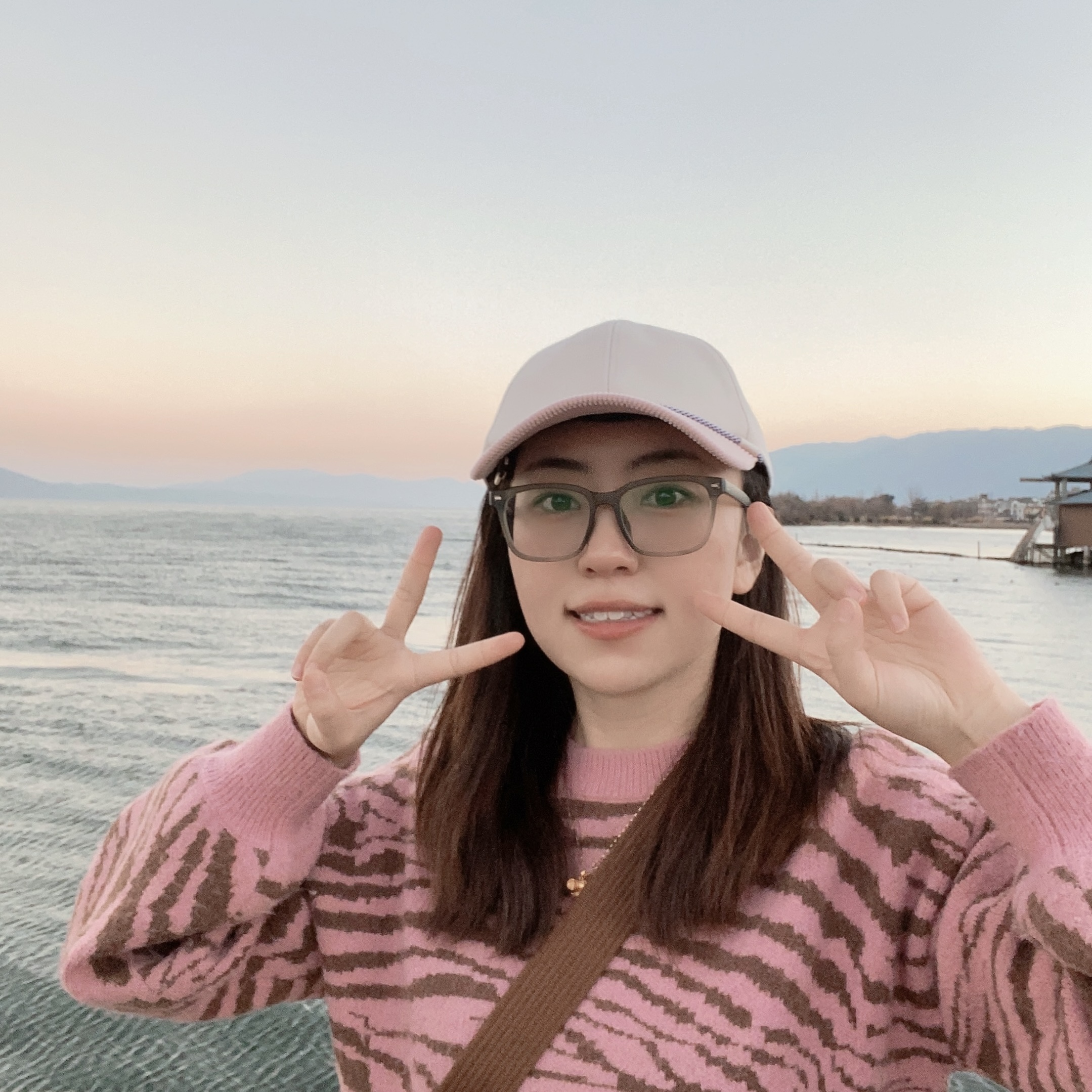}}]
{Yueru Luo} is currently pursuing the Ph.D. degree at The Chinese University of Hong Kong, Shenzhen. Her research interests include 3D computer vision, autonomous-driving perception, 3D lane detection, scene topology reasoning and scene understanding in autonomous-driving environments. Her recent work focuses on large-scale 4D geometry foundation models and vision-language models.
\end{IEEEbiography}

\vspace{-4mm}

\begin{IEEEbiography}[{\includegraphics[width=1in,height=1.25in,clip,keepaspectratio]{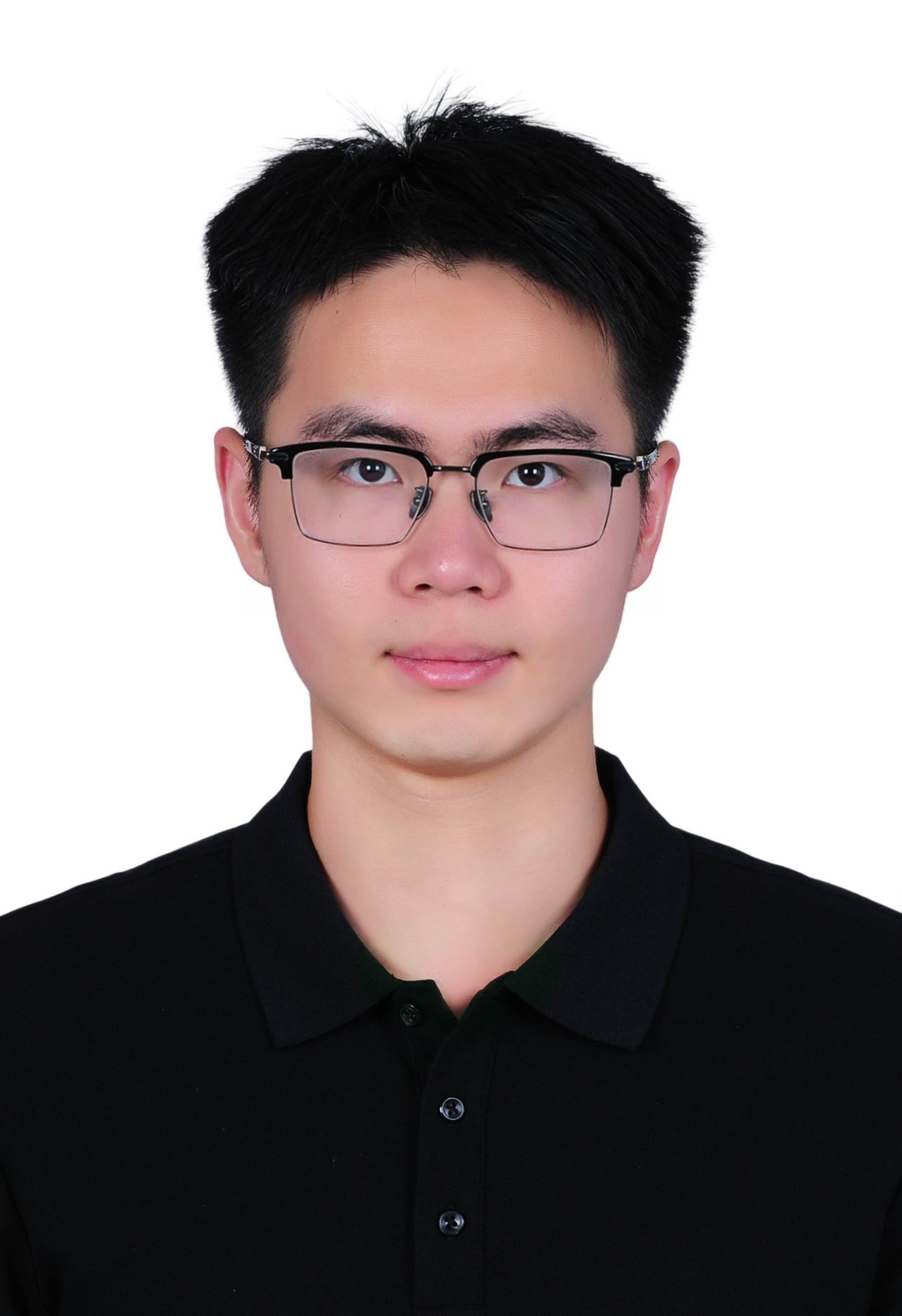}}]
{Han Zhang} is currently pursuing the M.Phil. degree at The Hong Kong University of Science and Technology (Guangzhou). His research interests include 3D computer vision, embodied intelligence, and spatial perception. His recent research focuses on visual navigation and 3D spatial understanding for embodied agents.
\end{IEEEbiography}

\vspace{-4mm}

\begin{IEEEbiography}[{\includegraphics[width=1in,height=1.25in,clip,keepaspectratio]{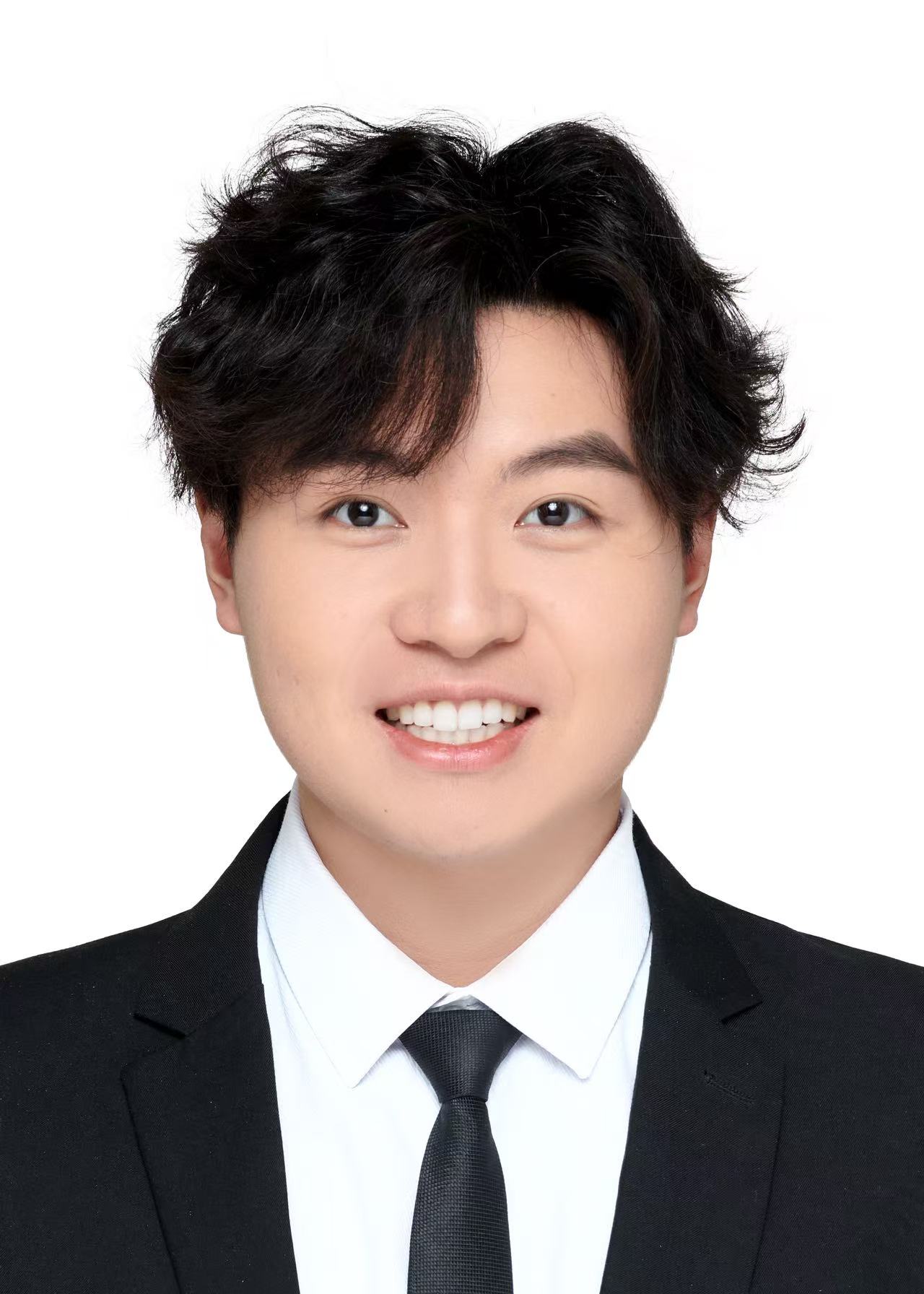}}]
{Zeyu Jiang} is currently pursuing the Ph.D. degree at The Hong Kong University of Science and Technology (Guangzhou). His research interests include 3D Computer Vision, Visual SLAM, 3D Representation and Embodied Navigation. His recent work focuses on physical agent operation system for embodied navigation.
\end{IEEEbiography}

\vspace{-4mm}

\begin{IEEEbiography}[{\includegraphics[width=1in,height=1.25in,clip,keepaspectratio]{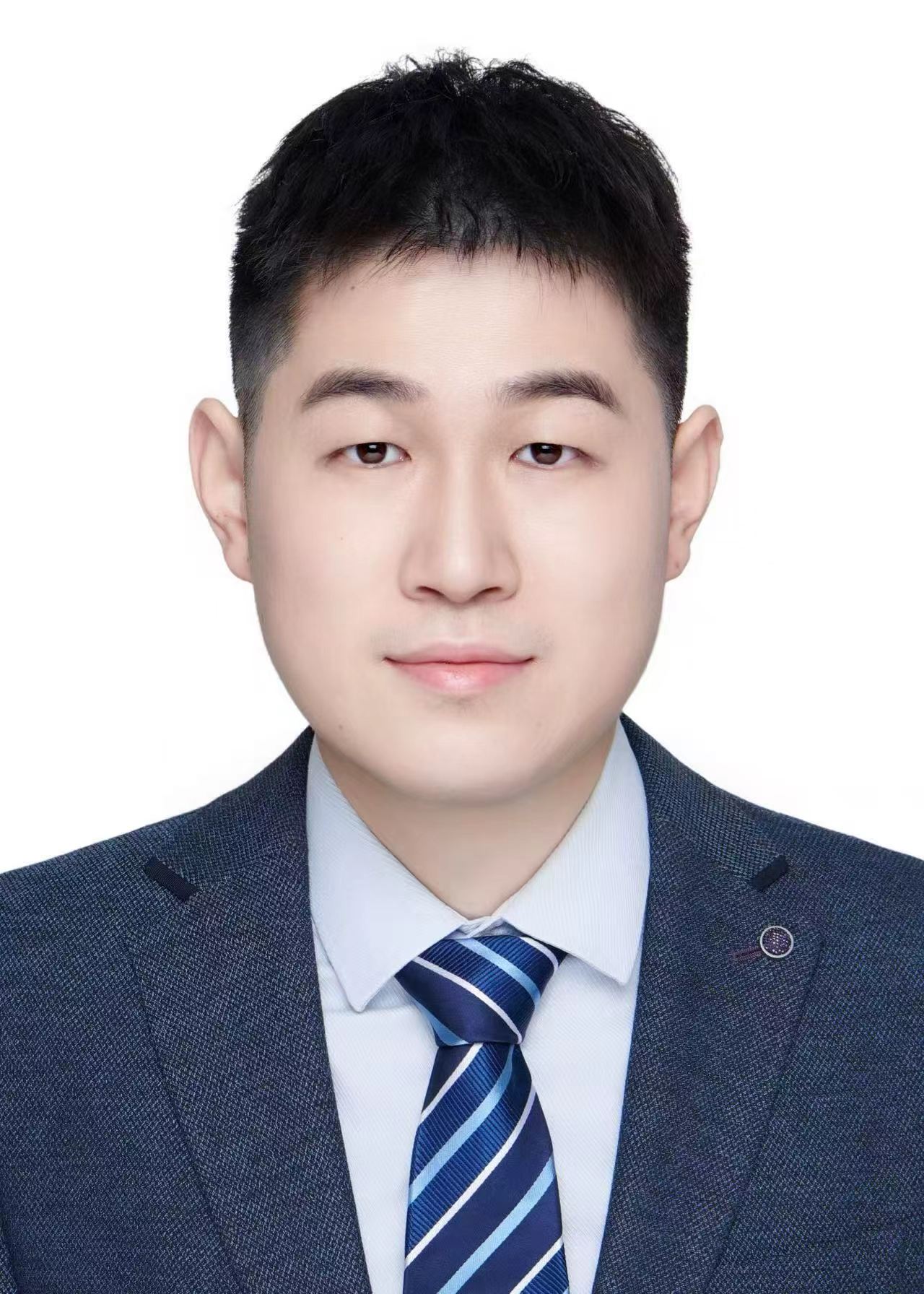}}]
{Changhao Chen} obtained his Ph.D. degree at University of Oxford (UK), M.Eng. degree at the National University of Defense Technology (China), and B.Eng. degree at Tongji University (China). Now he is an Assistant Professor at the Thrust of Intelligent Transportation and Thrust of Artificial Intelligence, the Hong Kong University of Science and Technology (Guangzhou). His research interest lies in robotics, embodied AI and cyber-physical systems.
\end{IEEEbiography}

\end{document}